\documentclass[10pt,journal,compsoc]{IEEEtran}
\ifCLASSOPTIONcompsoc
  \usepackage[nocompress]{cite}
\else
  \usepackage{cite}
\fi
\ifCLASSINFOpdf
\else
  Online Image Vectorizern (dvipsone, dvipdf, if not using dvips). graphicx
\fi

\usepackage{subfigure}
\usepackage{times}
\usepackage{epsfig}
\usepackage{hyperref}
\usepackage{graphicx}
\usepackage{amsmath}
\usepackage{amssymb}
\usepackage{multirow}
\usepackage{tabularx}
\usepackage{booktabs}

\usepackage{amsmath, bm}
\usepackage{algorithm,algpseudocode}
\usepackage{xcolor,colortbl}
\usepackage{color, colortbl}
\definecolor{LightCyan}{rgb}{0.88,1,1}
\definecolor{LightRed}{rgb}{1,0.88,0.95}

\usepackage{pifont}

\newcommand{\ie}{\textit{i.e. }}
\newcommand{\eg}{\textit{e.g. }}
\newcommand{\vs}{\textit{vs. }}
\newcommand{\etal}{\textit{et al. }}

\begin{document}
%
\title{ST3D++: Denoised Self-training for Unsupervised Domain Adaptation on 3D Object Detection}
%
%
%
%

\author{
Jihan Yang,~Shaoshuai Shi,~Zhe Wang,~Hongsheng Li,~Xiaojuan Qi 
\IEEEcompsocitemizethanks{\IEEEcompsocthanksitem Jihan Yang and Xiaojuan Qi are with the Department of Electrical and Electronic Engineering at The University of Hong Kong, Hong Kong. Shaoshuai Shi and Hongsheng Li are with the Department of Electronic Engineering at The Chinese University of Hong Kong, Hong Kong. Zhe Wang is with SenseTime Research.
\IEEEcompsocthanksitem {Email: jhyang@eee.hku.hk, xjqi@eee.hku.hk}
}
}

\IEEEtitleabstractindextext{%
\begin{abstract}
In this paper, we present a 
{self-training} method, named ST3D++, with a holistic pseudo label denoising pipeline for unsupervised domain adaptation on 3D object detection.
ST3D++ aims at reducing noise in pseudo label generation as well as alleviating the negative impacts of noisy pseudo labels on model training. 
First, ST3D++ pre-trains the 3D object detector on the labeled source domain with random object scaling (ROS) which is designed to reduce target domain pseudo label noise arising from object scale bias of the source domain. 
Then, the detector is progressively improved through alternating between generating pseudo labels
and training the object detector with pseudo-labeled target domain data.
Here, we equip the pseudo label generation process with a hybrid quality-aware triplet memory to improve the quality and stability of generated pseudo labels. Meanwhile, in the model training stage, we propose a source data assisted training strategy and a curriculum data augmentation policy to effectively rectify noisy gradient directions and avoid model over-fitting to noisy pseudo labeled data. 
These specific designs enable the detector to be trained on meticulously refined pseudo labeled target data with denoised training signals, and thus effectively facilitate adapting an object detector to a target domain without requiring annotations.
Finally, our method is assessed on four 3D benchmark datasets ({\ie} Waymo, KITTI, Lyft, and nuScenes) for three common categories ({\ie} car, pedestrian and bicycle). ST3D++ achieves state-of-the-art performance on all evaluated settings, outperforming the corresponding baseline by a large margin (\eg 9.6\% $\sim$ 38.16\% on Waymo $\rightarrow$ KITTI in terms of AP$_{\text{3D}}$), and even surpasses the fully supervised oracle results on the KITTI 3D object detection benchmark with target prior. Code will be available. 


\end{abstract}

}

\maketitle

\IEEEdisplaynontitleabstractindextext

%
\IEEEpeerreviewmaketitle

\IEEEraisesectionheading{\section{Introduction}\label{sec:introduction}}

%
%
%
%
\IEEEPARstart{3}{D} object detection aims to categorize and localize objects from 3D sensor data ({\eg} LiDAR point clouds) with many applications in autonomous driving, robotics, virtual reality, to name a few.
Recently, this field has obtained remarkable advancements~\cite{yan2018second,lang2019pointpillars,shi2019pointrcnn,shi2019points,shi2020pv,shi2021pv} driven by deep neural networks and large-scale human-annotated datasets~\cite{Geiger2012KITTI,sun2020scalability,caesar2020nuscenes,lyft2019}.

However, 3D detectors developed on one specific domain ({\ie} training / source domain) might not generalize well to novel testing domains ({\ie} target domains) due to unavoidable domain-shifts arising from different types of 3D depth sensors, weather conditions and geographical locations, etc. For instance, a 3D detector trained on data collected in USA cities with Waymo LiDAR ({\ie} Waymo dataset~\cite{sun2020scalability}) suffers from a dramatic performance drop (of over $45\%$)~\cite{wang2020train} when evaluated on data from European cities captured by Velodyne LiDAR ({\ie} KITTI dataset~\cite{Geiger2012KITTI}). Though collecting more training data from different domains could alleviate this problem, it unfortunately might be infeasible given various real-world scenarios and enormous costs for 3D annotation. Therefore, approaches to effectively adapting 3D detectors trained on a labeled source domain to a new unlabeled target domain is highly demanded in practical applications. This task is also known as unsupervised domain adaptation (UDA) for 3D object detection.

In contrast to intensive studies on UDA in the 2D setting \cite{ganin2014unsupervised,long2015learning,hoffman2016fcns,chen2018domain,saito2019strong,ge2019mutual,ge2020self}, 
few efforts~\cite{wang2020train} have been made to explore UDA for 3D object detection. Meanwhile, the fundamental differences in data structures and network architectures render UDA approaches for image tasks not readily applicable to this problem.  
For domain adaptation on 3D detection, while promising results have been obtained in~\cite{wang2020train}, the method requires object size statistics of the target domain, and its efficacy largely depends on data distributions.
\begin{figure*}[t]
    \centering
    \includegraphics[width=1\linewidth]{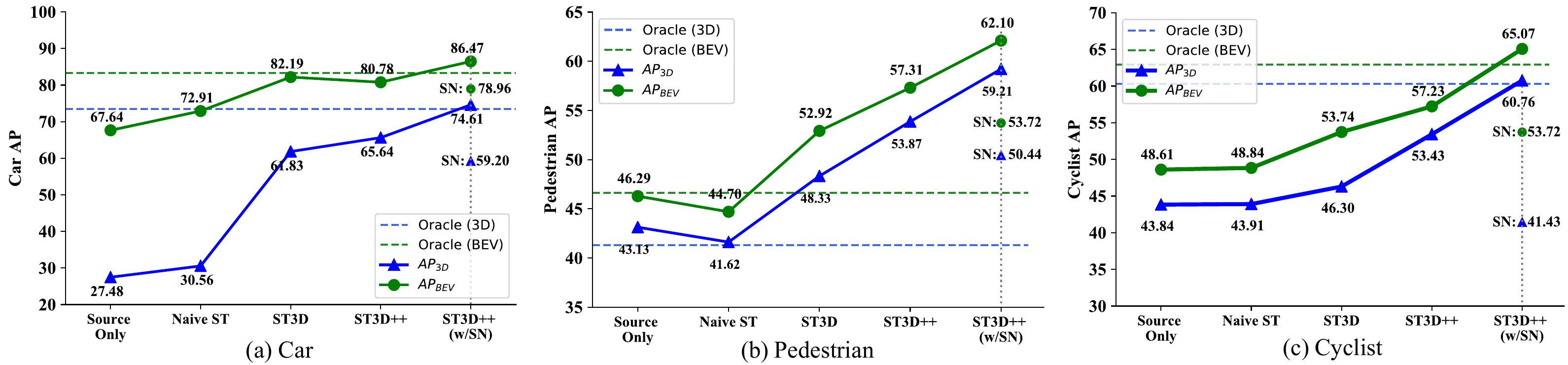}
    \caption{Performance of ST3D++ on the Waymo $\rightarrow$ KITTI task using SECOND-IoU \cite{yan2018second} for car, pedestrian and cyclist. Our ST3D++ is compared with other unsupervised (\ie source only, naive ST), weakly-supervised (\ie SN \cite{wang2020train}) and fully supervised (\ie oracle) approaches. Dashed line denotes the performance of Second-IoU on labeled target data in the fully supervised setting.}
    \label{fig:performance_comp}
\end{figure*}

Recently, self-training has emerged as a simple and effective technique for UDA, attaining state-of-the-art performance on many image understanding tasks~\cite{zhang2020label,zou2019confidence,khodabandeh2019robust}. 
Self-training starts from pre-training a model on source labeled data and further iterating between pseudo label generation and model training on unlabeled target data until convergence is achieved. It formulates the task of UDA as a supervised learning problem on the target domain with pseudo labels, which explicitly closes the domain gaps. 
Despite of encouraging results in image tasks, our study illustrates that naive self-training~\cite{xie2020self} does not work well in UDA for 3D object detection as shown in Fig.~\ref{fig:performance_comp} (``source only'' {\vs} ``naive ST'').


The major obstacle for self-training on domain adaptive 3D object detection lies in severe pseudo label noise, such as imprecise object localization ({\ie} orientated 3D bounding box) and incorrect object categories, which are yet largely overlooked in the naive self-training pipeline. 
The noise is the compound effect of the domain gap between source and target domain data 
and the capability of the 3D object detector (\eg systematic errors of the 3D detector). These noisy pseudo labels will not only misguide the direction of model optimization but also make errors accumulate during iterative pseudo label generation and model training, leading to inferior performance.

In this paper, we propose a holistic denoised self-training pipeline for UDA on 3D object detection, namely ST3D++, which simultaneously reduces target domain pseudo label noise and mitigates the negative impacts of noisy pseudo labels on model training.

First, in model pre-training on source domain labeled data, we develop \textit{random object scaling} (ROS), a simple 3D object augmentation technique, which randomly scales the 3D objects to overcome the bias in object scale of the source domain and effectively reduces pseudo label noise from biased source domain data.
Second, for pseudo label generation in iterative self-training, we develop a \textit{hybrid quality-aware triplet memory} which encompasses a hybrid box scoring criterion to assess the quality of object localization and categorization, a triplet box partition scheme to avoid assigning pseudo labels to inconclusive examples, and a memory updating strategy, integrating historical pseudo labels via ensemble and voting, to reduce pseudo label noise and instability.
Finally, in the model training process, we design a \textit{source-assisted self-denoised} (SASD) training method with separate source and target batch normalization, which fully leverages the advantage of clean and diverse source annotations to rectify the direction of gradients as well as address negative impacts of joint optimization on source and target domain data with domain shifts.
Meanwhile, \textit{curriculum data augmentation} (CDA) is developed for pseudo-labeled target data to guarantee effective learning at the beginning and gradually simulate hard examples through progressively increasing the intensity of augmentation. CDA also prevents the model from overfitting to easy examples -- pseudo-labeled data with high confidence -- and thus improves model's capabilities.


Experimental results on four 3D object detection datasets KITTI~\cite{Geiger2012KITTI}, Waymo~\cite{sun2020scalability}, nuScenes~\cite{caesar2020nuscenes}, and Lyft~\cite{lyft2019} for three common categories including car, pedestrian and cyclist, demonstrate the effectiveness of our approach. The performance gaps between source only results and fully supervised oracle results are closed by a large percentage. Besides, we outperform existing approaches~\cite{wang2020train,saltori2020sf,you2021exploiting} by a notable margin (around 13\% $\sim$ 17\%) based on the same setup. It's also noteworthy that our approach even outperforms the oracle results for all categories on the Waymo $\rightarrow$ KITTI setting when further combined with target statistics~\cite{wang2020train} as shown in Fig.~\ref{fig:performance_comp}. 

\vspace{0.1cm}
\noindent
\textbf{Different from our conference paper:} This manuscript signiﬁcantly improves the conference version~\cite{yang2021st3d}: 
(i) We extend domain adaptive 3D detection to multiple categories (\ie, car,  pedestrian and cyclist) which is the first attempt on four popular 3D detection datasets: Waymo~\cite{sun2020scalability}, nuScenes~\cite{caesar2020nuscenes}, KITTI~\cite{Geiger2012KITTI} and Lyft~\cite{lyft2019} as far as we know.
(ii) We conduct more analysis on pseudo label noise for self-training and present a holistic pseudo label denoised self-training pipeline, which addresses pseudo label noise in a systematic manner from model pre-training, pseudo label generation to model optimization.
(iii) We improve the quality-aware criterion to account for both the localization quality and the classification accuracy. 
(iv) We propose a source-assisted training strategy where source examples are leveraged in the self-training stage to rectify incorrect gradient directions from noisy pseudo labels and provide more diverse patterns. Besides, the domain-specific normalization is incorporated to avoid the negative impacts of mixing source and target data in source-assisted joint optimization.
(v)  We carry out a study on the quality of pseudo labels on 3D object detection where five quantitative indicators are proposed, and conduct more analysis on how the proposed strategies improve pseudo label qualities.   
(vi) We conduct extensive experiments on four datasets for three categories. The proposed ST3D++ outperforms ST3D~\cite{yang2021st3d} over all categories on all adaptation settings as well as other existing UDA methods, SF-UDA$^{3D}$~\cite{saltori2020sf}, Dreaming~\cite{you2021exploiting} and MLC-Net~\cite{luo2021consistency} by a large margin.
(vii) We explore to leverage temporal information to further improve ST3D++ through fused sequential point frames, which also mitigates the point density gaps across domains.

\section{Related Work}
\label{sec:related work}
\noindent
\textbf{3D Object Detection from Point Clouds} aims to localize and classify 3D objects from point clouds, 
which is a challenging task due to the irregularity and sparsity of 3D point clouds.
Some previous works \cite{chen2017multi, ku2018joint,yang2018pixor} proposes to resolve this task by previous 2D detection methods directly via mapping the irregular 3D point clouds to 2D bird eye's view grids. 
Another line of research \cite{yan2018second, zhou2018voxelnet,shi2019points,he2020structure, shi2020pv} adopts 3D convolutional networks to learn 3D features from voxelized point clouds, and the extracted 3D feature volumes are also further compressed to bird-view feature maps as above.  
Recently, point-based approaches \cite{shi2019pointrcnn, yang2019std} propose to directly generate 3D proposals from raw point clouds by adopting PointNet++~\cite{qi2017pointnet++} to extract point-wise features. There are also some other methods \cite{qi2018frustum, wang2019frustum} that utilize 2D images for generating 2D box proposals which are further employed to crop the object-level point clouds to produce 3D bounding boxes. 
In our work, we adopt SECOND~\cite{yan2018second}, PointRCNN~\cite{shi2019pointrcnn} and PV-RCNN~\cite{shi2020pv} as our 3D object detectors.

\vspace{0.2cm}
\noindent
\textbf{Unsupervised Domain Adaptation}
targets at obtaining a robust model which can generalize well to target domains with only labeled source examples and unlabeled target data. 
Previous works \cite{long2015learning,long2018conditional} explore domain-invariant feature learning by minimizing the Maximum Mean Discrepancy \cite{ben2010impossibility}. Inspired by GANs~\cite{goodfellow2014generative}, adversarial learning was employed to align feature distributions across different domains on image classification \cite{ganin2014unsupervised,ganin2016domain}, semantic segmentation \cite{hoffman2016fcns,tsai2018learning} and object detection \cite{chen2018domain,saito2019strong} tasks. Besides, Benefited from the development of unpaired image to image translation \cite{zhu2017unpaired}, some methods~\cite{hoffman2017cycada,zhang2018fully,gong2019dlow} proposed to mitigate the domain gap on pixel-level by translating images across domains. Another line of approaches \cite{saito2017asymmetric,zou2018unsupervised,khodabandeh2019robust,cai2019exploring} leverage self-training~\cite{lee2013pseudo} to generate pseudo labels for unlabeled target domains. Saito \etal \cite{saito2018maximum} adopted a two branch classifier to reduce the $\mathcal{H} \Delta \mathcal{H}$ discrepancy. \cite{li2018adaptive,wang2019transferable,chang2019domain} alleviated the domain shift on batch normalization layers by modulating the statistics in BN layer before evaluation or specializing parameters of BN domain by domain.   
\cite{supancic2013self,choi2019pseudo,chitta2018adaptive} employed curriculum learning~\cite{bengio2009curriculum} and separated examples by their difficulties to realize local sample-level curriculum. Xu \etal \cite{Xu_2019_ICCV} proposed a progressive feature-norm enlarging method to reduce the domain gap. \cite{liu2019transferable,yang2020adversarial} injected feature perturbations to obtain a robust classifier through adversarial training. 

On par with the developments on domain adaptation for image recognition tasks, some recent works also aim to address the domain shift on point clouds for shape classification~\cite{qin2019pointdan} and semantic segmentation~\cite{wu2018squeezeseg,yi2020complete,jaritz2020xmuda}. However, despite intensive studies on the 3D object detection task~\cite{zhou2018voxelnet, shi2019pointrcnn,yan2018second,shi2019points, yang2019std, shi2020pv}, only few approaches have been proposed to solve UDA for 3D object detection. Wang \etal proposed SN \cite{wang2020train} to normalize the object size of the source domain leveraging the object statistics of the target domain to close the size-level domain gap. Though the performance has been improved, the method needs the target statistics information, and its effectiveness depends on the source and target data distributions.
SF-UDA$^{3D}$~\cite{saltori2020sf} and Dreaming~\cite{you2021exploiting} leveraged the extra tracker and time consistency regularization along target point cloud sequence to generate better pseudo labels for self-training.  
SRDAN~\cite{zhang2021srdan} employed scale-aware and range-aware feature adversarial alignment manners to match the distribution between two domains, which might suffer from stability and convergence issues. MLC-Net~\cite{luo2021consistency} employed the mean-teacher paradigm to address the geometric mismatch between source and target domains.
In contrast, the proposed ST3D++ tailors a self-denoising framework to simultaneously close content and point distribution gaps across domains and achieve superior performance on all three categories of four adaptation tasks with neither prior target object statistics nor extra computation.


\section{Analysis of Pseudo Label Noise in Domain Adaptive 3D Object Detection}
Self-training~\cite{lee2013pseudo} obtains remarkable progress on unsupervised domain adaptation~\cite{saito2017asymmetric,zou2018unsupervised,khodabandeh2019robust,cai2019exploring} by alternating between pseudo-labeling target data and updating the model, which explicitly closes the domain gap by re-formulating the UDA problem as a target domain supervised learning problem with pseudo labels. If pseudo labels are ideally perfect, the problem will be close to supervised learning on the target domain. 
Hence, the quality of pseudo labels is the determining factor for successful domain adaptation. 
In the following, we will analyze pseudo label noise in 3D object detection.

\begin{figure}[t]
	\centering
	\includegraphics[width=\linewidth]{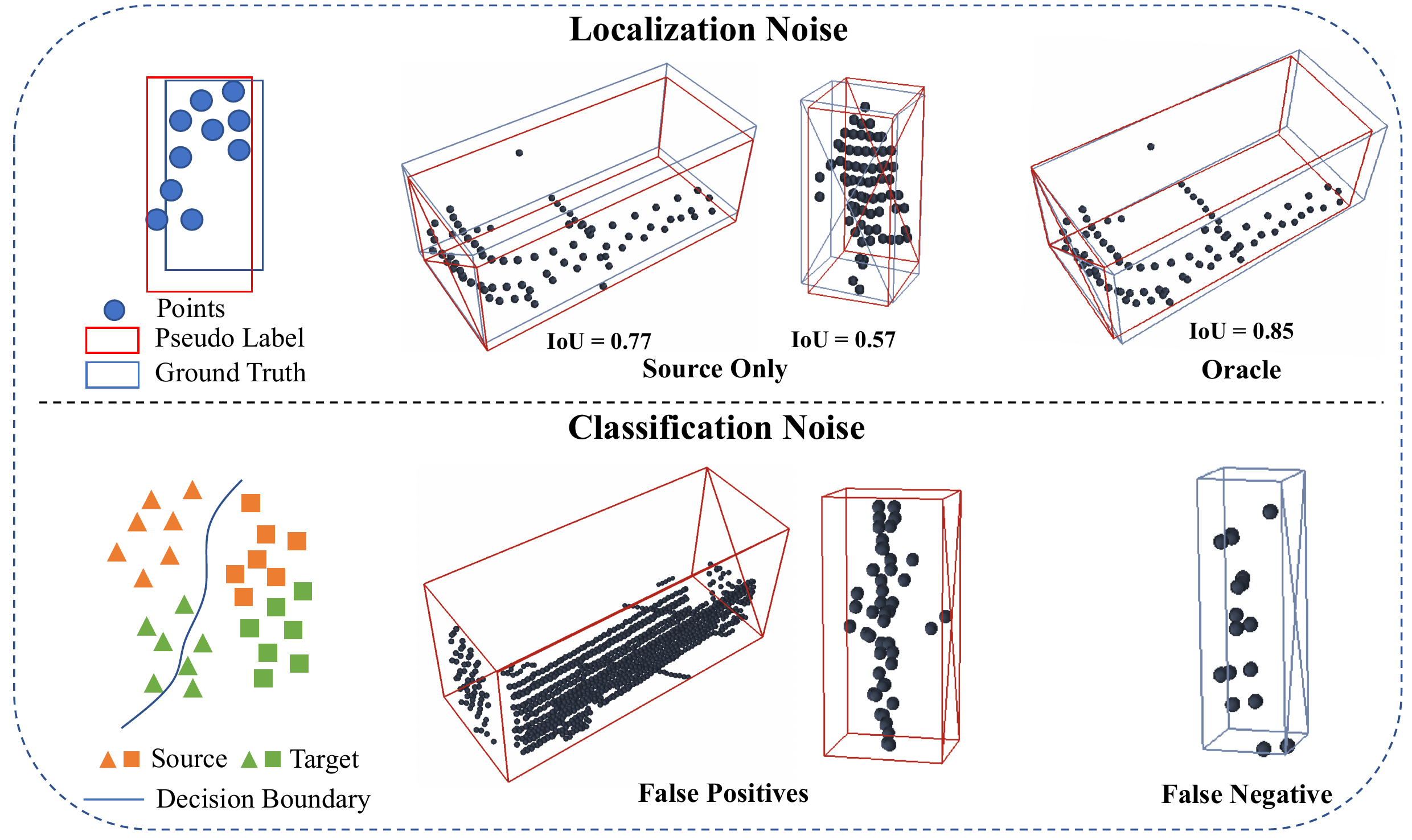}
	\caption{Examples of two types self-training noise (\textcolor{red}{red} and \textcolor{blue}{blue} boxes stand for predicted pseudo labels and GTs, respectively). Upper row:
	"localization noise" consists of pseudo labels that locate GTs well but suffers from regression errors to their corresponding GTs. Bottom row: ``classification noise'' mainly includes false-positive and false-negative pseudo labels (\ie missed GTs). }
    \label{fig:noise}
\end{figure}

As 3D object detection requires to jointly categorize and localize objects, the pseudo label noise could be divided into localization noise and classification noise as shown in Fig.~\ref{fig:noise}.  
For these two types of noise, their causes including domain shifts and model capabilities, and their negative impacts on domain adaptive 3D object detection with self-training are elaborated as below.


\vspace{0.1cm}
\noindent
\textbf{Localization noise} -- low-quality pseudo labeled bounding boxes -- is unavoidable in pseudo labeling the target domain data.
On the one hand, the object detector has to estimate the 3D bounding boxes with orientations based on incomplete point cloud, which is an ill-posed and challenging problem itself with ambiguities (see Oracle prediction in the upper row of Fig.~\ref{fig:noise}).
And different source and target point cloud patterns further intensify the difficulties for the model to produce accurate bounding box predictions on the target domain data. As shown in the upper row of Fig.~\ref{fig:noise}, pseudo label of car predicted by source only achieves much lower IoU with GTs compared to the pseudo label obtained by the oracle model (\eg {Source only: $0.77$, Oracle: $0.85$}) due to few noise points.
On the other hand, object sizes and annotation rules vary in different datasets (\eg cars in Waymo dataset around 0.9 meters longer than cars in KITTI on average), which will arise negative transfers in 3D bounding boxes from the source domain to the target domain, generating inaccurate 3D target domain pseudo-labeled 3D boxes.

\begin{figure*}[t]
	\centering
	\includegraphics[width=\linewidth]{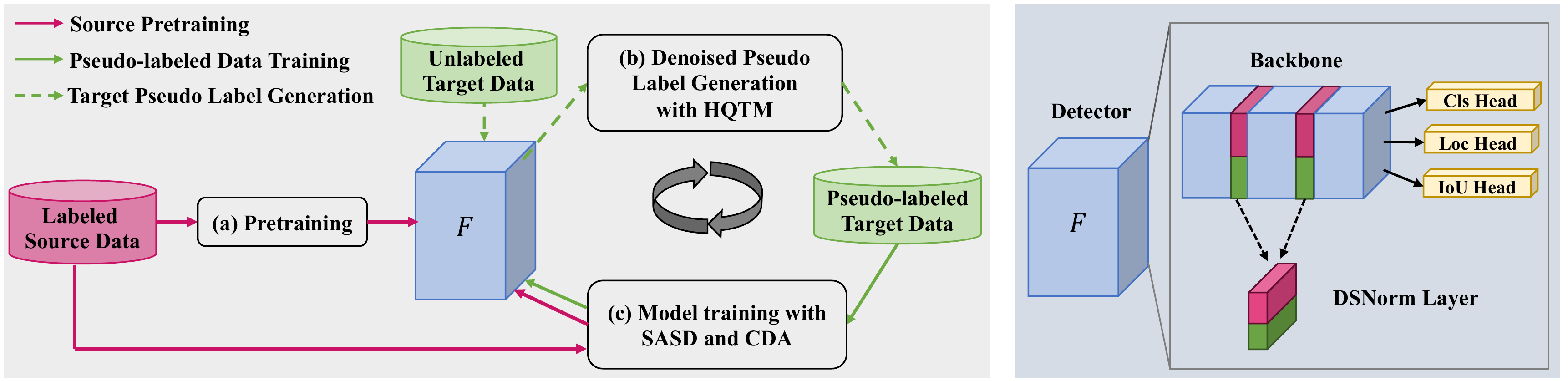}
	\caption{Left: An overview of ST3D++ framework, which includes three stages: (a) model pre-training with ROS; (b) Denoised pseudo label generation with HQTM and (c) model training with SASD and CDA. Object detector $F$ iterates between the second and the third stages until convergence. Right: An architecture overview of the object detector. Our ST3D++ is a model-agnostic framework. Any one-stage and two-stage detectors with classification head, localization head and IoU head can be incorporated into our framework.}
    \label{subfig:framework_p1}
\end{figure*}

\vspace{0.1cm}
\noindent
\textbf{Classification noise} 
refers to false positives (\ie, backgrounds or other incorrect categories, that are misclassified and assigned to pseudo labeled bounding boxes) and false negatives (\ie missed objects) as shown in the bottom row of Fig.~\ref{fig:noise}.
The main causes lie in the following folds: $(i)$ different point cloud patterns in source and target domains confuse the object detector, leading to incorrect class predictions; and $(ii)$ detector's capability is insufficient to distinguish backgrounds and different object categories; and  $(iii)$ sub-optimal criteria to generate pseudo categories. 

\vspace{0.1cm}
\noindent
\textbf{Pseudo label noise}, including localization and classification noise in 3D object detection, is unavoidable and yet detrimental to the self-training process if not properly handled.
First, the noisy pseudo labeled data will produce imprecise gradients which will guide the updating of model weights in an incorrect direction.
Then, the negative impacts will be amplified as the updated model will be further used to produce more noisy pseudo labeled data, making the training process fall into a vicious circle of error accumulation.

\begin{algorithm}[h]
	\small
	\caption{Pipeline of our ST3D++. 
	}
	\renewcommand{\algorithmicensure}{ \textbf{Output:}} 
	\label{algo:our_pipeline}
    \begin{algorithmic}[1]
		\Require 
		Source domain labeled data $\{(P^s_i, L^s_i)\}^{n_s}_{i=1}$, and target domain unlabeled data $\{P^t_i\}_{i=1}^{n_t}$.
		\Ensure The object detection model for target domain.
		
		\State {Pre-train} the object detector $F$ on $\{(P^s_i, L^s_i)\}^{n_s}_{i=1}$ with ROS as detailed in Sec.~\ref{sec:pretrain}. \label{algo:line_ros}
		\State Utilize the current model to generate {raw object proposals} $[B^{t}_i]_k$ for every sample $P_i^t$, where $k$ is the current number of times for pseudo label generation.
		\State Generate quality-aware pseudo labels $[\hat{L}^{t}_i]_k$ by triplet box partition given $[B^{t}_i]_k$ in Sec.~\ref{sec:triplet}. \label{algo:line_triplet}
		\State Update the memory ({\ie} pseudo labels) $[M_i^{t}]_k$ given pseudo labels $[\hat{L}^{t}_i]_k$ from the detection model and historical pseudo labels $[M_i^{t}]_{k-1}$ $([M_i^{t}]_0 = \emptyset)$ in the memory with memory ensemble-and-voting (MEV) as elaborated in Sec.~\ref{sec:memoryensemble}.
		The memory $\{[M^{t}_i]_k\}_{i=1}^{n_t}$ consists of pseudo labels for all unlabeled examples.
		\State Train the model on $\{P_i^t, [M_i^{t}]_k\}_{i=1}^{n_t} $ and $\{(P^s_i, L^s_i)\}^{n_s}_{i=1}$ with SASD and CDA for several epochs as detailed in Sec.~\ref{sec:self_training}. \label{algo:line_CDA}
		\State Go back to Line 2 until convergence.
	\end{algorithmic}
\end{algorithm}

\section{ST3D++}
\subsection{Overview}
Our goal is to adapt a 3D object detector trained on source labeled data $\{(P^s_i, L^s_i)\}^{n_s}_{i=1}$ with $n_s$ samples to unlabeled target data $\{P^t_i\}_{i=1}^{n_t}$ with $n_t$ samples. Here, $P^s_i$ and $L^s_i$ represent the $i$-th source input point cloud and its corresponding label.  
$L^s_i$ contains the category and 3D bounding box information for each object in the $i$-th point cloud, and each box is parameterized by its center $(c_x, c_y, c_z)$, size $(l, w, h)$ and heading angle $\theta$. Similarly, $P^t_i$ denotes the $i$-th unlabeled target point cloud.

\begin{figure*}[t]
	\centering
	\includegraphics[width=\linewidth]{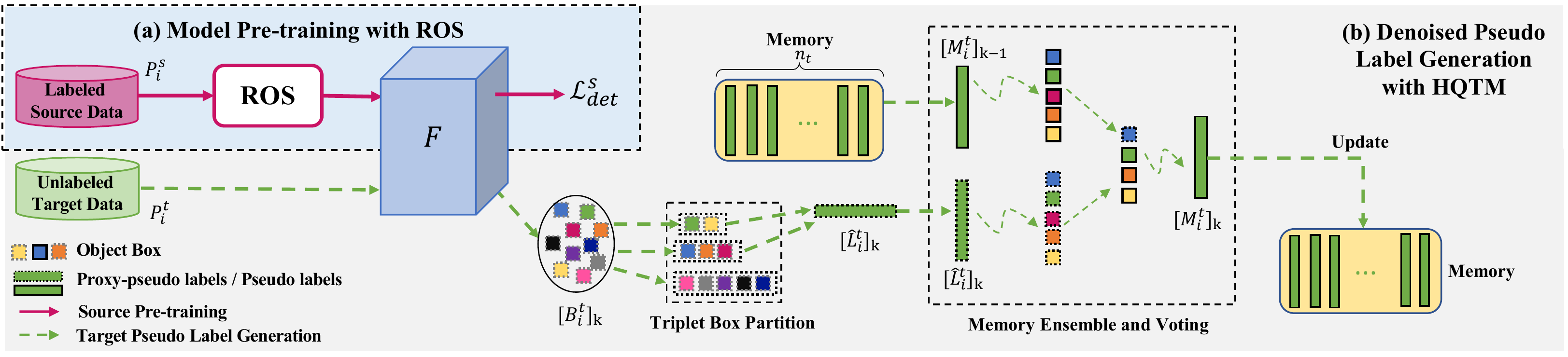}
	\caption{Overview of stage (a) and (b). (a) Pre-train the object detector $F$ with random object scaling (ROS) as data augmentation in source domain to mitigate noise owing to source object-size bias. (b) Generate high-quality and consistent pseudo labels on target unlabeled data with our hybrid quality-aware triplet memory (HQTM).}
    \label{subfig:framework_p2}
\end{figure*}


We present ST3D++, a self-training framework with a holistic pseudo-label denoising pipeline, to adapt the 3D detector trained on the source domain to the target domain. 
Our ST3D++ tackles the noise issues in pseudo labeling for 3D object detection through 
reducing the above noises to generate high-quality pseudo labels
and mitigating negative impacts of noisy labels during the model training.   
An overview of our approach is shown in Fig.~\ref{subfig:framework_p1} and described in Algorithm~\ref{algo:our_pipeline}.
First, the object detector is pre-trained on the source labeled domain with  random object scaling (ROS) (see Fig.~\ref{subfig:framework_p2} (a)) which mitigates source domain bias and facilitates obtaining a robust object detector for pseudo label generation.
Then, the object detector is progressively improved by alternating
between generating pseudo labels on the target data, where hybrid quality-aware triplet memory (HQTM) (see Fig.~\ref{subfig:framework_p2} (b)) is designed to denoise the pseudo labeled data and enforce pseudo label consistency through training, and fine-tuning the object detector on pseudo labeled target data, where a source-assisted self-training (SASD) method is proposed to rectify imprecise gradient directions from noisy labels and a curriculum data augmentation (CDA) (see Fig.~\ref{subfig:framework_p3}) strategy is incorporated to avoid model overfitting to pseudo-labeled easy samples.  

The rest of this section is organized as follows. Sec.~\ref{sec:pretrain} presents model pre-training with ROS. Then, Sec.~\ref{sec:memory_bank} details the denoised pseudo label generation process with HQTM. Finally, Sec.~\ref{sec:self_training} elaborates the model training process with SASD and CDA.

\subsection{Model Pre-training with Random Object Scaling}\label{sec:pretrain}

ST3D++ starts from training a 3D object detector on labeled source data  $\{(P^s_i, L^s_i)\}^{n_s}_{i=1}$. The pre-trained model learns how to perform 3D detection on source labeled data and is further adopted to initialize object predictions for the target domain unlabeled data. 

 \noindent
 \textbf{Observations.}~
However, despite the useful knowledge, the pre-trained detector also learns the bias from the source data, such as object size and point densities.
Among them, the bias in object size has direct negative impacts on 3D object detection, and results in noisy pseudo-labeled target bounding boxes with incorrect sizes. This is also in line with the findings in~\cite{wang2020train}. To mitigate the issue, we propose a very simple and yet effective object-wise augmentation strategy, {\ie} \textit{random object scaling} (ROS), fully leveraging the high degree of freedom of 3D spaces. 

\vspace{0.1cm}
\noindent
\textbf{Random Object Scaling.}~
Given an annotated 3D bounding box with size $(l, w, h)$, center $(c_x, c_y, c_z)$ and heading angle $\theta$, ROS scales the box in the length, width and height dimensions with random scale factors $(r_l, r_w, r_h)$ through transforming the points inside the box. 
We denote the points inside the box as $\{p_i\}_{i=1}^{n_p}$ with a total of $n_p$ points, and the coordinate of $p_i$ is represented as $(p_i^x, p_i^y, p_i^z)$.
First, we transform the points to the object-centric coordinate system of the box along its length, width and height dimensions via
\begin{equation}
\begin{aligned}
(p_i^l, p_i^w, p_i^h) &= (p_i^x - c_x, p_i^y - c_y, p_i^z - c_z) \cdot R, \\
 R &=
	\begin{bmatrix}
	\cos \theta & -\sin \theta & 0\\
	\sin \theta & \cos \theta & 0\\
	0 & 0 & 1\\
	\end{bmatrix},
    \end{aligned}
\end{equation}
where $\cdot$ is matrix multiplication.
Second, to derive the scaled object, the point coordinates inside the box are scaled to be $(r_lp_i^l, r_wp_i^w, r_hp_i^h)$ with object size $(r_ll, r_ww, r_hh)$.
Third, to derive the augmented data $\{p_i^{\text{aug}}\}_{i=1}^{n_p}$, the points inside the scaled box are transformed back to the ego-car coordinate system and shifted to the center $(c_x, c_y, c_z)$ as
\begin{equation}
p_i^\text{aug} = (r_lp_i^l, r_wp_i^w, r_hp_i^h)\cdot R^T + (c_x, c_y, c_z).
\end{equation}
Albeit simple, ROS effectively simulates objects with diverse object sizes to address the size bias and hence facilitates to train de-biased object detectors that produce more accurate initial pseudo boxes on target domain for subsequent self-training. 

\subsection{De-noised Pseudo Label Generation with Hybrid Quality-aware Triplet Memory}\label{sec:memory_bank}

With the trained detector, the next step is to generate pseudo labels for the unlabeled target data.
Given the target sample $P_i^t$, the output $B_i^t$ of the object detector is a group of predicted boxes containing category labels, confidence scores, regressed box sizes, box centers and heading angles, where non-maximum-suppression (NMS) has already been conducted to remove duplicated detections. 
For clarity, we denote $B_i^t$ as outputs from the object detector for the $i$-th sample.

\begin{figure}[t]
    \centering
    \includegraphics[width=\linewidth]{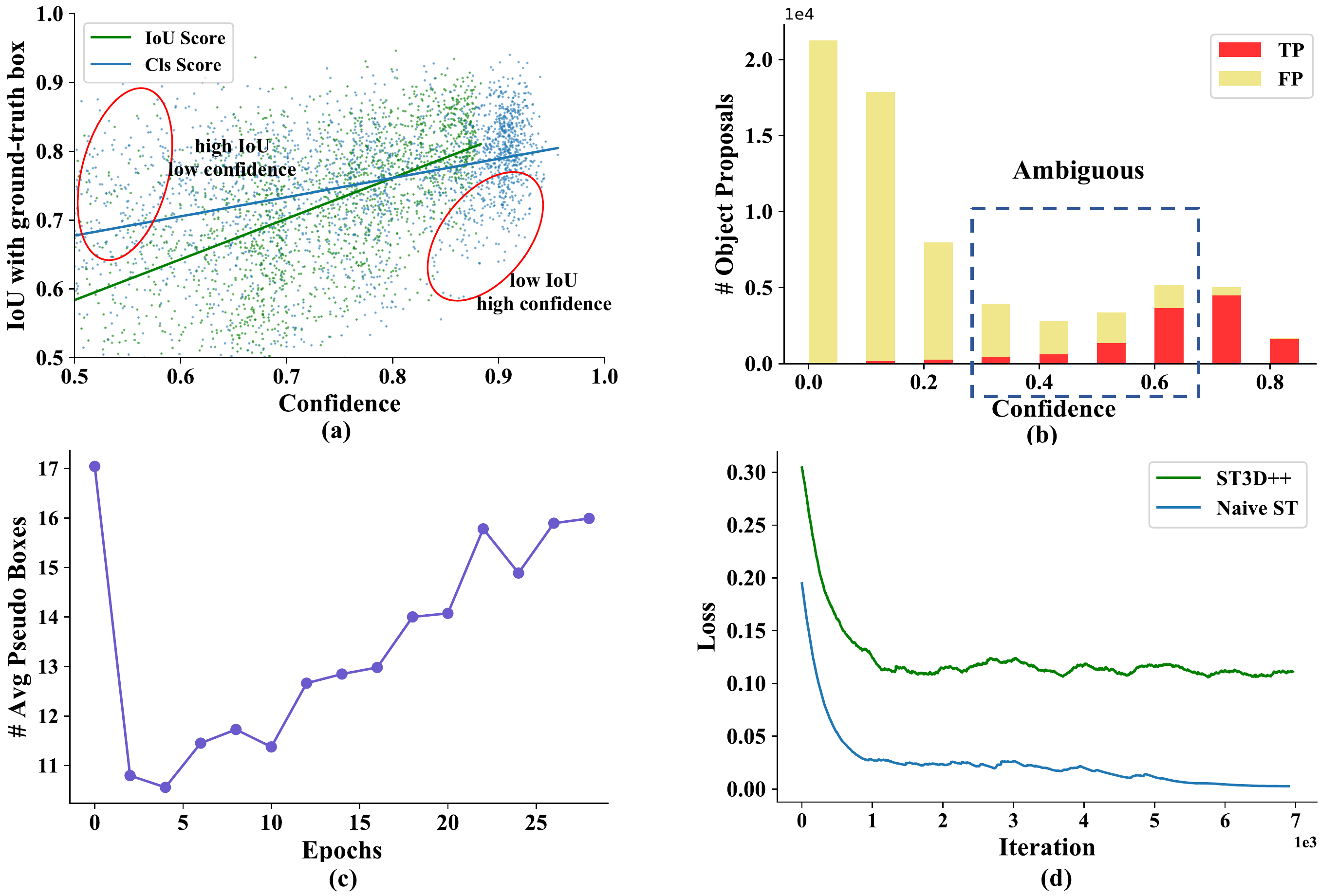}
    \caption{(a) Correlation between confidence value and box IoU with ground-truth (b) Lots of proposal boxes with medium confidence may be assigned with false labels.
    (c) The average number of pseudo boxes fluctuates at different epochs in the pseudo label generation stage. (d) Training loss curve comparison between naive ST and our ST3D++ with CDA.}
    \label{fig:motivations}
\end{figure}

 \noindent
 \textbf{Observations.}
Different from classification and segmentation tasks, 3D object detection needs to jointly consider the classification and localization information, which poses great challenges for high-quality pseudo label generation. 
First, the confidence of object category prediction may not reflect the precision of localization as shown by the blue line in Fig.~\ref{fig:motivations} (a). 
Second, the fraction of false labels is greatly increased in confidence score intervals with medium values as illustrated in Fig.~\ref{fig:motivations} (b). 
Third, model fluctuations induce
inconsistent pseudo labels as demonstrated in Fig.~\ref{fig:motivations}~(c). The above factors will undoubtedly introduce negative impacts on pseudo-labeled objects, leading to noisy supervisory information and instability for model. Hence, reducing noise of pseudo labels 
in the pseudo label generation stage is essential for self-training.

To this end, we design \textit{hybrid quality-aware triplet memory} (HQTM) to parse noisy object predictions $\{B_i^t\}_{i=1}^{n_t}$ into high-quality pseudo labels $\{M_i^t\}_{i=1}^{n_t}$, which are cached into the memory at each stage $k$ for self-training.
To obtain high-quality $\{M_i^t\}_{i=1}^{n_t}$, HQTM includes two major denoising components tailored to 3D object detection.  
Given noisy predictions $\{[B_i^t]_k\}_{i=1}^{n_t}$, the first denoising component incorporates a hybrid criterion (see Sec.~\ref{sec:triplet}) to assess quality of each predictions regarding localization and classification, and a triplet partition scheme to produce proxy-pseudo labels\footnote{To differentiate the intermediate pseudo labels $\{[\hat{L}_i^t]_k\}_{i=1}^{n_t}$ from the object detector and pseudo labels $\{[M_i^t]_{k-1}\}_{i=1}^{n_t}$  in the memory,  we call $\{[\hat{L}_i^t]_k\}_{i=1}^{n_t}$ ``\textit{proxy-pseudo label}''.} denoted as $\{[\hat{L}_i^t]_k\}_{i=1}^{n_t}$ (see Sec.~\ref{sec:tripletbox}), aiming to avoid assigning labels to object predictions with ambiguous confidence. 
Then, to further enhance pseudo label qualities, the second denoising component (see Sec.~\ref{sec:memoryensemble}) combines the proxy pseudo labels $\{[\hat{L}_i^t]_k\}_{i=1}^{n_t}$ and historical pseudo labels $\{[M_i^t]_{k-1}\}_{i=1}^{n_t}$ in the memory through the elaborately designed memory ensemble and voting strategies. The overall pipeline of this procedure is demonstrated in Fig.~\ref{subfig:framework_p2} (b).




\begin{table*}[]
\renewcommand\arraystretch{1.1}
    \centering
    \caption{TP ratio of object predictions produced by IoU or classification score on different score intervals, respectively. For pedestrian, classification score provides more accurate pseudo labels (\ie a higher TP ratio) for object predictions with high confidence (\eg confidence larger than 0.7). ``-'' indicates no predictions fall into this score interval.}
    \begin{small}
    \setlength{\tabcolsep}{5.1mm}{
    \begin{tabular}{c|c|c|c|c|c|c}
    \bottomrule[1pt]
       \multirow{2}{*}{Category} & \multirow{2}{*}{Criterion} & \multicolumn{5}{c}{Score Interval} \\
       \cline{3-7}
        & & 0.4 - 0.5 & 0.5 - 0.6 & 0.6 - 0.7 & 0.7  - 0.8 & 0.8 - 0.9 \\
       \hline
       \multirow{2}{*}{Car} & IoU & 0.18 & 0.34 & 0.63 & 0.87 & 0.93 \\
       \cline{2-7}
        & Cls & 0.40 & 0.51 & 0.66 & 0.80 & 0.92 \\
        \hline
       \multirow{2}{*}{Pedestrian}  & IoU & 0.11 & 0.38 & 0.80 & 0.67 & - \\
       \cline{2-7}
         & Cls & 0.38 & 0.55 & 0.76 & 0.90 &0.94\\
    \toprule[0.8pt]   
    \end{tabular}
    }
    \end{small}
    
    \label{tab:tp_ratios}
\end{table*}

\subsubsection{Hybrid Quality-aware Criterion for Scoring} \label{sec:triplet}

Although classification confidence is a widely adopted criterion to measure the quality of predictions~\cite{saito2017asymmetric,zou2018unsupervised,khodabandeh2019robust} in self-training, it fails to reflect the localization quality in 3D object detection as shown in Fig.~\ref{fig:motivations} (a).
To simultaneously evaluate the quality of localization and classification, we propose an IoU-based criterion which is further integrated with the confidence-based criterion to assess prediction qualities detailed as below.

\noindent 
{\bf IoU-based Criterion for Scoring.}
To directly assess the localization quality of pseudo labels, we propose to augment the original object detection model with a lightweight IoU regression head. 
Specifically, given the feature derived from RoI pooling, we append two fully connected layers to directly predict the 3D box IoU between RoIs and their ground truths (GTs) or pseudo labels. 
A sigmoid function is adopted to map the output into range $[0,1]$.
During model training, the IoU branch is optimized using a binary cross-entropy loss as
\begin{equation}
    \label{eq:iou_loss}
	\mathcal{L}_{\text{iou}}=-\hat{u}\log u - (1 - \hat{u})\log (1-u),
\end{equation}
where $u$ is the predicted IoU and $\hat{u}$ is the IoU between the ground truth (or pseudo label) box and the predicted 3D box.
%

\noindent 
{\bf Hybrid Criterion for Better Scoring. } 
For object categories that are easily distinguishable from backgrounds ({\eg} ``cars''), the IoU score not only correlates well with localization quality (see Fig.~\ref{fig:motivations} (a)), but also generates more true positives (TPs) in confident interval (\ie score larger than $0.7$ in Table~\ref{tab:tp_ratios}) if adopted as a criterion. 
However, the classification score enjoys an obvious superiority in categories that are easily confused with backgrounds ({\eg} pedestrians similar to background ``trees'' and `` poles''), {\ie} the classification score obtains pseudo labels with higher TP ratios in the high-confident regime compared to the IoU score (\ie confidence larger than 0.7) in Table~\ref{tab:tp_ratios}. 


To take the best of both worlds, we propose an effective hybrid quality-aware criterion which integrates classification confidence and IoU scores in a weighted manner as 
\begin{equation}
    o = \phi p + (1-\phi) u, \label{eq: weight}
\end{equation}
where $o$ is the final criterion for each 3D box, $p$ is the classification score of object predictions and $\phi$ is the trade-off parameters between $p$ and $u$ ($\phi$ is set to 0 $\sim$ 0.5 across different tasks).


\subsubsection{Triplet Box Partition to Avoid Ambiguous Samples}\label{sec:tripletbox}
Now, we are equipped with a hybrid quality assessment criterion to assess $[B^t_i]_k$ (for the $i$-th sample at stage $k$) from the detector after NMS.
Here, to avoid assigning labels to ambiguous examples which introduce noise, we present a triplet box partition scheme to obtain the proxy-pseudo labels $[\hat{L}_i^t]_k$.
Given an object box $b$ from $[B^t_i]_k$ with the final criterion $o_b$, we create a  margin $[T_{\text{neg}}, T_\text{pos}]$ to ignore boxes with score $o_b$ inside this margin, preventing them from contributing to training, as follows:
\begin{align}
\text{state}_b\! =\! \!\left\{
\begin{tabular}{@{}l@{}}
$\text{Positive} \text{ (Store to } [\hat{L}_i^t]_k \text{)}, \  {T_{\text{pos}} \leq o_b}, $\\
$\text{Ignored} \text{ (Store to } [\hat{L}_i^t]_k \text{)} , \  {T_{\text{neg}} \leq o_b < T_{\text{pos}}},$\\
$\text{Negative} \text{ (Discard)} , \ \ \ \ \ \ \ \ \ \ \ \ \  {o_b < T_{\text{neg}}}.$
\end{tabular}
\right. 
\end{align}
If $\text{state}_b$ is positive, $b$ will be cached into $[\hat{L}_i^t]_k$ as a positive sample with its state, category label, pseudo box and confidence. Similarly, ignored boxes will also be incorporated into the $[\hat{L}_i^t]_k$ to identify regions that should be ignored during model training due to their high uncertainty. Box $b$ with negative $\text{state}_b$ will be discarded, corresponding to backgrounds.

Our triplet box partition scheme reduces pseudo label noise caused by ambiguous boxes and thus ensures the quality of pseudo-labeled boxes. 
It also shares a similar spirit with curriculum learning~\cite{bengio2009curriculum}, where confident easy boxes are learned at earlier stages and ambiguous hard examples are handled later after the model has been trained well enough to distinguish these examples.
\subsubsection{Memory Update and Pseudo Label Generation}~\label{sec:memoryensemble}
\noindent
Here, we combine proxy-pseudo labels $\{[\hat{L}_i^t]_k\}_{i=1}^{n_t}$ at stage $k$ and the historical pseudo labels $\{[M_i^{t}]_{k-1}\}_{i=1}^{n_t}$ $([M_i^{t}]_0 = \emptyset)$ in the memory via memory ensemble and voting to leverage historical pseudo labels to perform pseudo label denoising and obtain consistency-regularized pseudo labels at stage $k$. 
The outputs are the updated pseudo labels $\{[M_i^{t}]_{k}\}_{i=1}^{n_t}$ that  will serve as supervisions for subsequent model training procedures.
During this memory update process, each pseudo box $b$ from $[\hat{L}_i^t]_k$ and $[M_i^{t}]_{k-1}$ has three attributes $(o_b, \text{state}_b, \text{cnt}_b)$, which are the hybrid quality-aware confidence score, state (positive or ignored)  and an unmatched memory counter (UMC) (for memory voting), respectively.
We assume that $[\hat{L}_i^t]_k$ contains $n_l$ boxes denoted as $[\hat{L}_i^t]_k = \{(o_l, \text{state}_l, \text{cnt}_l)_j^k\}_{j=1}^{n_l}$ and  $[M_i^t]_{k-1}$ has $n_m$ boxes represented as $[M_i^t]_{k-1} = \{(u_m, \text{state}_m, \text{cnt}_m)_j^{k-1}\}_{j=1}^{n_m}$.


\begin{figure}[t]
    \centering
    \includegraphics[width=1\linewidth]{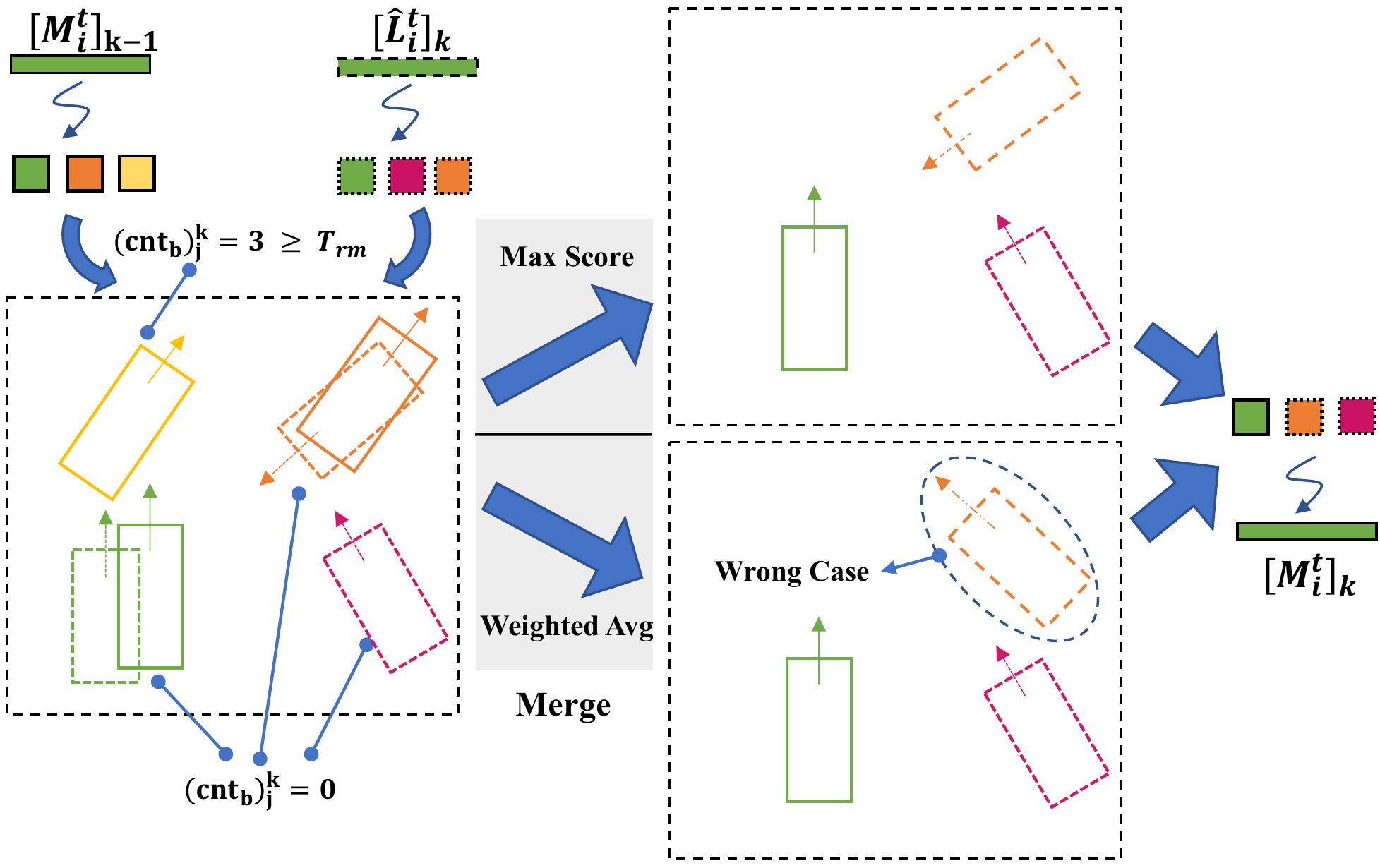}
    \caption{An instance of memory ensemble and voting (MEV). Given proxy-pseudo labels $[\hat{L}_i^t]_k$ and historical pseudo labels $[M_i^t]_{k-1}$, MEV automatically matches and merges boxes while ignoring or discarding successively unmatched boxes (\eg yellow and red boxes are unmatched, but considering their $(\text{cnt}_b)^k_j$), only the yellow box will be discarded for thrice successively unmatching). The weighted average boxes merging strategy could produce wrong final boxes for boxes with very different heading angles. Best view in color.}
    \label{fig:memory_ensemble}
    \vspace{-0.2cm}
\end{figure}
 
\vspace{0.1cm}
\noindent
\textbf{Memory Ensemble.}
Instead of directly replacing $[M_i^t]_{k-1}$ with the latest proxy-pseudo labels $[\hat{L}_i^t]_k$, we propose the memory ensemble operation to combine $[M_i^t]_{k-1}$ and $[\hat{L}_i^t]_k$, which denoises the pseudo labels through consistency checking considering historical pseudo labels. 

The memory ensemble operation matches two object boxes with similar locations, sizes and angles from $[M_i^t]_{k-1}$ and $[\hat{L}_i^t]_k$, and merges them to produce a new object box. 
By default, we adopt the \textbf{consistency ensemble} strategy for box matching. 
Specifically, it calculates the pair-wise 3D IoU matrix $A = \{a_{jv}\} \in \mathbb{R}^{n_m \times n_l}$ between each box in $[M_i^t]_{k-1}$ and each box in $[\hat{L}_i^t]_k$. 
For the $j$-th object box in $[M_i^t]_{k-1}$, its matched box index $\hat{j}$ in $[\hat{L}_i^t]_k$ is derived by,
\begin{equation}
    \hat{j} = \text{argmax}_j~ (a_{jv}), ~{v=1, \cdots, n_l}.
\end{equation}
Note that if $a_{j\hat{j}} < 0.1$, we denote each of these two paired boxes as unmatched boxes that will be further processed by the memory voting operation.

We assume the successfully matched pair-wise object boxes as  $(o_l, \text{state}_l, \text{cnt}_l)_{\hat{j}}^{k}$ and 
$(o_m, \text{state}_m, \text{cnt}_m)_{j}^{k-1}$. 
They are further merged to cache the pseudo labeled box with a higher confidence value into the $[M_i^{t}]_{k}$ and update its corresponding attributes as
\begin{equation}
\label{eq:me_state_update}
(o_m, \text{state}_m, 0)_j^{k} = \!\left\{\!
\begin{array}{l}\!
\!(o_l, \text{state}_l, \text{cnt}_l)_{\hat{j}}^{k} , \ \ \ \ \text{if} ~{o_m \!\leq\! o_l}, \\\!
\!(o_m, \text{state}_m, \text{cnt}_m)_j^{k-1}, \text{otherwise}, 
\end{array}
\right.
\end{equation}
%
Here, we adopt to choose box instead of a weighted combination since weighted combination has the potential to produce an unreasonable final box if the matched boxes have very different heading angles (see Fig.~\ref{fig:memory_ensemble} ``wrong case''). 
Similarly, 3D Auto Labeling~\cite{qi2021offboard} also observes that selecting box candidates with highest confidence is more optimal.

In addition, we also explore other two memory ensemble variants for box ensembling. For the first variant \textbf{NMS ensemble}, it is an intuitive solution to match and merge two types of boxes 
by removing the duplicated boxes based on IoU. 
It directly removes matched boxes with lower confidence scores.
Specifically, we concatenate historical pseudo labels and current proxy-pseudo labels to $[\tilde{M}^{t}_i]_{k} = \{[M^{t}_i]_{k-1}, [\hat{L}^t_i]_k\}$ as well as their corresponding confidence scores to $\tilde{o}^k = \{o^{k-1}_m, o^k_l \}$ for each target sample $P^t_i$. Then, we obtain the final pseudo boxes $[M^{t}_i]_{k}$ and corresponding confidence score $o^k_m$ by applying NMS with an IoU threshold at $0.1$ as
\begin{equation}
    [M^{t}_i]_{k}, o^k_m = \text{NMS}([\tilde{M}^{t}_i]_{k},\  \tilde{o}^k).
\end{equation}
For the second variant \textbf{bipartite ensemble}, it employs optimal bipartite matching~\cite{carion2020end} to pair historical pseudo labels $[M_i^{t}]_{k-1}$ and current proxy-pseudo labels $[\hat{L}^t_i]_k$ and then follow consistency ensemble to process matched pairs. Concretely, we assume that there are $n_m$ and $n_l$ boxes for $[M_i^{t}]_{k-1}$ and $[\hat{L}^t_i]_k$ separately. Then, we search a permutation of $n_m$ elements $\kappa \in \mathfrak{S}_{n_m}$ with the lowest cost as
\begin{equation}
    \hat{\kappa}=\underset{\kappa \in \mathfrak{S}_{n_m}}{\arg \min } \sum_{j}^{n_m} \mathcal{L}_{\operatorname{match}}\left(b_{j}, b_{\kappa(j)}\right),
\end{equation}
where the matching cost $\mathcal{L}_{\operatorname{match}}$ is the $-\text{IoU}$ between the matched boxes. Notice that the matched box pairs with IoU lower than 0.1 would still be regarded as unmatched. 

\vspace{0.1cm}
\noindent
\textbf{Memory Voting.}
The memory ensemble operation can effectively select better matched pseudo boxes. 
However, it cannot handle the unmatched pseudo boxes from either $[M_i^t]_{k-1}$ or $[\hat{L}_i^t]_k$.
As the unmatched boxes often contain both false positive boxes and true positive boxes, either caching them into the memory or discarding them all is sub-optimal.  
To address the above problem, we propose a novel memory voting approach, which leverages history information of unmatched object boxes to robustly determine their status (\textbf{cache}, \textbf{discard} or \textbf{ignore}).
%
%
For the $j$-th {unmatched} pseudo boxes $b$ from $[M_i^t]_{k-1}$ or $[\hat{L}_i^t]_k$, its unmatched memory counter (UMC) $(\text{cnt}_b)_j^{k}$ will be updated as follows: 
\begin{equation}
(\text{cnt}_b)^k_{j} = \left\{
\begin{array}{lll}
0 &, & \text{if } b \in [\hat{L}_i^t]_k,  \\
(\text{cnt}_b)_j^{k-1} + 1 & , & \text{if }  b \in [M_i^t]_{k-1},\\
\end{array}
\right.
\end{equation}
We update the UMC for unmatched boxes in $[M_i^t]_{k-1}$ by adding $1$ and initialize  the UMC of the newly generated boxes in  $[\hat{L}_i^t]_k$ as 0.
The UMC records the successfully unmatched times of a box, which are combined with two thresholds $T_{\text{ign}}$ and $T_{\text{rm}}$ 
($T_{\text{ign}}=2$ and $T_{\text{rm}}=3$ by default) 
to select the subsequent operation for unmatched boxes as
\begin{equation}
\left\{\!
\begin{array}{llc}\!
\text{Discard} &, & (\text{cnt}_b)^k_{j} \geq T_{\text{rm}} ,  \\\!
\text{Ignore}\ (\text{Store to} [M_i^{t}]_{k}) & , & T_{\text{ign}} \leq (\text{cnt}_b)^k_{j} < T_{\text{rm}} , \\
\!\text{Cache}\ (\text{Store to} [M_i^{t}]_{k})& , & (\text{cnt}_b)^k_{j} < T_{\text{ign}} .\\
\end{array}
\right.
\end{equation}
Benefited from our memory voting, we could generate more robust and consistent pseudo boxes by caching the occasionally unmatched box in the memory.


\begin{figure*}[t]
	\centering
	\includegraphics[width=\linewidth]{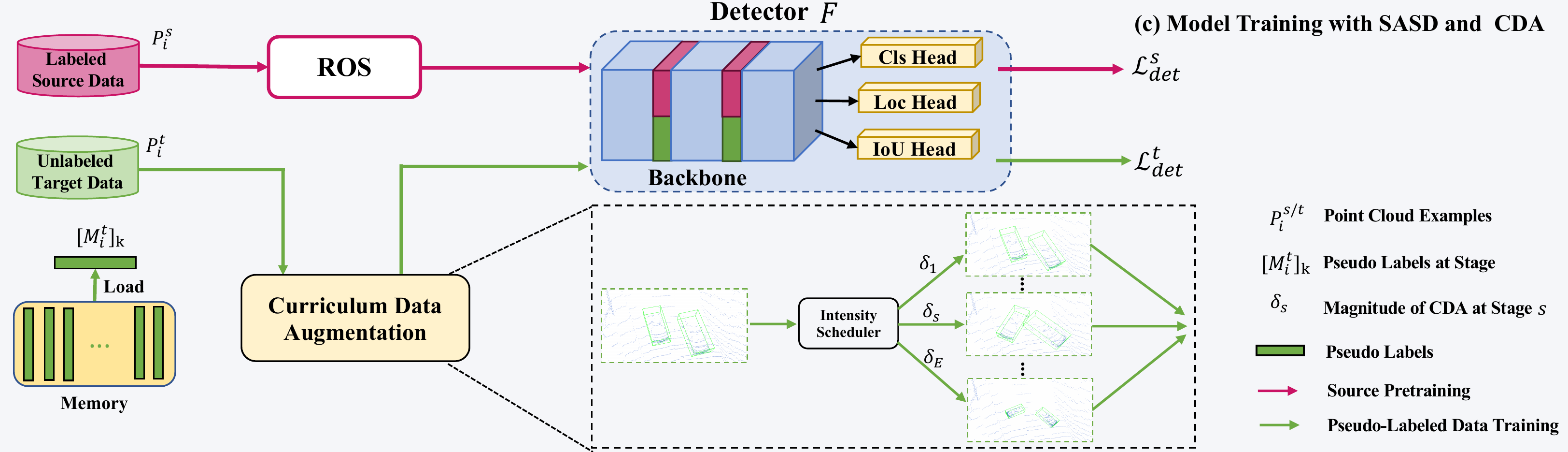}
	\caption{Pipeline overview of stage (c). Model is simultaneously optimized on both labeled source data and pseudo-labeled target data. Source-assisted self-denoising (SASD) is incorporated to rectify optimization directions and address normalization statistics shift, while curriculum data augmentation (CDA) prevents model overfitting to easy cases.}
    \label{subfig:framework_p3}
\end{figure*}

\subsection{Model training with SASD and CDA}\label{sec:self_training}

The proposed hybrid quality-aware triplet memory addresses pseudo label noise during the pseudo label generation stage and can produce high-quality and consistent pseudo labels $[M_i^{t}]_k$ for each point cloud $i$.
Now, the detection model can be trained on $\{P_i^t, [M_i^{t}]_k\}_{i=1}^{n_t}$ and $\{(P^s_i, L^s_i)\}^{n_s}_{i=1}$ as described in Algorithm~\ref{algo:our_pipeline} (Line~\ref{algo:line_CDA}) and Fig.~\ref{subfig:framework_p3}. 
In the following, we will elaborate on how to alleviate the negative impacts of noisy pseudo labeled data on model training using source-assisted self-denoised training (see Sec.~\ref{sec:sasd}) and avoid model over-fitting with curriculum data augmentation (see Sec.~\ref{sec:cda}).



\subsubsection{Source-assisted Self-denoised Training}
\label{sec:sasd}
Although we have attempted to mitigate pseudo label noise at model pre-training and pseudo label generation stages as discussed above, pseudo labels still can not be totally noise-free. 
The pseudo label noise will unfortunately disturb the direction of model updating, and errors will be accumulated during iterative self-training. Here, we propose source-assisted self-denoised (SASD) training to make the optimization be more noise-tolerant and ease error accumulations by joint optimization on source data and pseudo-labeled target data. However, simultaneously optimizing data from different domains could induce domain shifts and degrade model performance. To tackle this issue, we adopt domain specific normalization to avoid the negative impacts of joint optimization on data with different underlying distributions. Details of domain specific normalization and the joint training optimization objectives are elaborated as below. 


\vspace{0.1cm}
\noindent
\textbf{Domain Specific Normalization.}
Batch normalization (BN)~\cite{ioffe2015batch} is an extensively employed layer in deep neural networks as it can effectively accelerate convergence and boost model performance. 
Nevertheless, BN suffers from transferability issues when being applied in cross-domain data scenarios as source and target examples are drawn from different underlying distributions.
This is also observed in the adversarial training~\cite{xie2020adversarial} community for handling out-of-domain data. 
In this regard, we replace each BN layer with a very simple Domain Specific Normalization (DSNorm) layer, which disentangles the statistic estimation of different domains at the normalization layer.
During training, DSNorm calculates batch mean $\mu$ and variance $\sigma^2$ for each domain separately  as Eq.~\eqref{eq:bn}.
\begin{equation}
\begin{aligned}
    \mu_d = \frac{1}{mp}\sum_{i=1}^{m}\sum_{j=1}^{p}f_{ij}^d,\ \ \ \  {\sigma_d^{2}} = \frac{1}{mp} \sum_{i=1}^{m}\sum_{j=1}^{p}(f_{ij}^{d} - \mu_{d})^2, \label{eq:bn}
\end{aligned}
\end{equation}
where $d \in \{s, t\} $ is the domain indicator for source and target domain respectively, $m$ is the total number of domain specific samples in a mini-batch, $f$ is the input feature of one feature channel, and $p$ is the number of elements (\eg, pixels or points) in this feature channel.
Here, we ignore the channel index $c$ for simplicity, and the above process is performed on each channel separately. 
Meanwhile, since the transformation parameters $\gamma$ and $\beta$ are domain agnostic and transferable across domains, the two domains shares the learnable scale and shift parameters as Eq.~\eqref{eq:normalization}.
\begin{equation}
\begin{aligned}
    \hat{f}^{d}_i = \frac{f_i^{d} - \mu_{d}}{\sqrt{ {\sigma_d^{2}} + \epsilon}}, \ \ \ \ \ \ \   g^{d}_i = \gamma  \hat{f}^{d}_i + \beta, \label{eq:normalization}
\end{aligned}
\end{equation}
where $\hat{f}$ is the normalized value and $g$ is the transformed feature.
%

At the inference stage on the target domain, model predictions are obtained using target $\bar{\mu}_t$ and $\bar{\sigma}_t$ by moving average over batch mean ${\mu}_t$  and variance ${\sigma}_t$ during training .

\vspace{0.2cm}
\noindent
\textbf{Optimization Objective.}
The detection loss $\mathcal{L}_{\text{det}}$ on each domain consists of four loss terms as below,
\begin{equation}
    \mathcal{L}_{\text{det}} =  \mathcal{L}_{\text{cls}} + \alpha_1 \mathcal{L}_{\text{reg}} + \alpha_2 \mathcal{L}_{\text{dir}} + \mathcal{L}_{\text{iou}}.
\end{equation}
The anchor classification loss $\mathcal{L}_{\text{cls}}$ is calculated using focal loss~\cite{lin2017focal} with default parameters to balance the foreground and background classes, $\mathcal{L}_{\text{reg}}$ employs smooth L1 loss to regress 
the residuals of box parameters,
$\mathcal{L}_{\text{dir}}$ utilizes binary cross entropy loss to estimate the ambiguity of heading angle $\theta$ as in~\cite{yan2018second,shi2020pv}, and IoU estimation loss $\mathcal{L}_{\text{iou}}$ is formulated as discussed in Eq.~\eqref{eq:iou_loss}. 
In addition, trade-off factors $\alpha_1$ and $\alpha_2$ (2.0 and 0.2 by default) are set to balance regression loss and direction loss as in \cite{yan2018second}. 
%
Then, considering both domains, the overall optimization objective at the model training stage is
\begin{equation}
    \mathcal{L} = \lambda \mathcal{L}^{s}_{\text{det}} + \mathcal{L}^{t}_{\text{det}},
\end{equation}
where $\mathcal{L}_{\text{det}}^{s}$ and $\mathcal{L}_{\text{det}}^{t}$ are detection losses for source and target domains respectively with the trade-off parameter $\lambda$ (default set as 1.0).

\noindent
\textbf{Analysis.} Source-assisted self-denoised self-training has the following merits that help alleviate the negative impacts of pseudo label noise on model training and improve model's robustness. First, by leveraging noise-free labeled source data, the imprecise gradients due to noisy pseudo labels can be effectively rectified. 
Then, through joint optimization on source and target domain data, the model will be enforced to learn domain-invariant features and maintain discriminativeness on diverse patterns across different domains. 
Further, the source domain data can provide the model with challenging cases that are easily ignored or misclassified ({\eg} ``tree'', ``pole'', \vs ``pedestrian'') on pseudo labeled target data. Moreover, the simple domain specific normalization scheme can effectively address domain shift issues in cross domain joint optimization, and further improve model performance.

\begin{table*}[ht]
\renewcommand\arraystretch{1.1}
    \centering
    \caption{Dataset overview. Note that we use \textbf{version 1.0} of Waymo Open Dataset. * indicates that the information is obtained from \cite{wang2020train}. ${\dagger}$ means that we count this statistical information only on the validation set.}
    \begin{tabular}{l|c|c|c|c|c|c|c}
        \bottomrule[1pt]
        Dataset  & \# Beam Ways & Beam Angles & \# Points Per Scene$^{\dagger}$ &  \# Training Frames & \# Validation Frames & Location & \# Night/Rain \\
        \hline
        Waymo \cite{sun2020scalability} & 64-beam & [-18.0$^{\circ}$, 2.0$^{\circ}$]$^{*}$ & 160k & 158,081 & 39,987 & USA & Yes/Yes \\
        \hline
        KITTI \cite{Geiger2012KITTI} & 64-beam & [-23.6$^{\circ}$, 3.2$^{\circ}$] & 118k & 3,712 & 3,769 & Germany & No/No \\
        \hline
        Lyft \cite{lyft2019} & 64-beam & [-29.0$^{\circ}$, 5.0$^{\circ}$]$^{*}$ & 69k & 18,900 & 3,780 & USA & No/No \\
        \hline
        nuScenes \cite{caesar2020nuscenes} & 32-beam & [-30.0$^{\circ}$, 10.0$^{\circ}$] & 25k & 28,130 & 6,019 & USA and Singapore & Yes/Yes \\
        \toprule[0.8pt]
    \end{tabular}
    \label{tab:dataset_infos}
\end{table*}

\subsubsection{Curriculum Data Augmentation.} \label{sec:cda}
\label{sec:cda}
Our observation shows that most positive pseudo boxes are easy examples since they are generated from previous high-confident object predictions. 
Consequently, during training, model is prone to overfitting to these easy examples with low loss values (see Fig.~\ref{fig:motivations}~(d)), unable to further mine hard examples to improve the detector~\cite{bengio2009curriculum}.
To prevent model from being trapped by bad local minimal, strong data augmentations could be an alternative to generate diverse and potentially hard examples to improve the model.
However, this might confuse the learner and hence be harmful to model training at the initial stage.

\noindent 
\textbf{Curriculum Data Augmentation.}
Motivated by the above observation, we design a curriculum data augmentation (CDA) strategy to progressively increase the intensity $\delta$ of data augmentation and gradually generate increasingly harder examples to facilitate improving the model and ensure effective learning at the early stages. 

To progressively increase the intensity $\delta$ of data augmentations $\{D_i\}^{n_d}_{i=1}$ with $n_d$ types (\ie global points transformation and per-object points transformation), we design a multi-step intensity scheduler with initial intensity $\delta_0^i$ for the $i$-th data augmentation.
Specifically, we split the total training epochs into $E$ stages.
After each stage, the data augmentation intensity is multiplied by an enlarging ratio $\rho$ ($\rho > 1$, we use $\rho=1.2$ by default).
Thus, the data augmentation intensity for $i$-th data augmentation at stage $s$ ($1 \leq s \leq E$) is derived as $\delta_s^i = \delta_0^i {\rho}^{s-1}$.
Hence, the random sampling range of the $i$-th data augmentation could be calculated as follows:
\begin{equation}
\left\{
\begin{array}{llc}
[-\delta_s^i,  \delta_s^i]  &, & \  \text{if} \ D_i \text{ belongs to rotation},  \\

[1-\delta_s^i,  1+\delta_s^i] &, &\text{if} \ D_i \text{ belongs to scaling}.\\
\end{array}
\right.
\end{equation}

\noindent CDA enables the model to learn from challenging samples while making the difficulty of examples be within the capability of the learner during the whole training process.
Experiments in Table~\ref{tab:abl_aug} illustrate its effectiveness via the curriculum regime . 





\section{Experiments}

\subsection{Experimental Setup}
\label{sec:exp_setup}

\noindent
\textbf{Datasets.}~
We conduct experiments on four widely used LiDAR 3D object detection datasets: KITTI~\cite{Geiger2012KITTI}, Waymo~\cite{sun2020scalability}, nuScenes~\cite{caesar2020nuscenes}, and Lyft~\cite{lyft2019}. The statistics of the four datasets are summarized in Table~\ref{tab:dataset_infos}. The domain gaps across different datasets mainly lie in two folds: $(i)$ content gap (\eg object size, weather condition, etc.) caused by different data-capture locations and time and $(ii)$ point distribution gap owing to different LiDAR types (\eg number of beam ways, beam range, vertical inclination and horizontal azimuth of LiDAR).

\begin{table*}[htbp]
\renewcommand\arraystretch{1.15}
    \centering
    \caption{Results of four different adaptation tasks. We report average precision (AP) in bird's-eye view ($\text{AP}_{\text{BEV}}$) and 3D  ($\text{AP}_{\text{3D}}$) of the car, pedestrian and cyclist at IoU threshold as 0.7, 0.5 and 0.5 respectively. The reported AP is for the moderate case when KITTI dataset is the source domain, and is the overall result for other settings. Note that results of ST3D~\cite{yang2021st3d} on pedestrian and cyclist are reproduced since they are not given. We indicate the best adaptation result by \textbf{bold}. ${\ddagger}$ indicates we apply random world sampling as an extra data augmentation on the source domain.}
    \begin{small}
    \setlength{\tabcolsep}{5.1mm}{
        \begin{tabular}{c|c|c|c|c}
            \bottomrule[1pt]
            {Task} & {Method}  & {Car} & {Pedestrian} & {Cyclist} \\
            \hline
            \multirow{7}{*}{Waymo $\rightarrow$ KITTI} & Source Only  & 67.64 / 27.48 & 46.29 / 43.13 & 48.61 / 43.84 \\
            & SN \cite{wang2020train} & 78.96 / 59.20 & 53.72 / 50.44 & 44.61 / 41.43 \\
            \cline{2-5}
            & ST3D & 82.19 / 61.83 & 52.92 / 48.33 & 53.73 / 46.09 \\
            & ST3D (w/ SN) & {85.83} / {73.37} & 54.74 / 51.92 & 56.19 / 53.00 \\
            \cline{2-5}
            & ST3D++ &  80.78 / 65.64 & 57.13 / 53.87 & 57.23 / 53.43  \\
            & ST3D++ (w/ SN) & \textbf{86.47} / \textbf{74.61} & \textbf{62.10} / \textbf{59.21} & \textbf{65.07} / \textbf{60.76} \\
            \cline{2-5}
            & Oracle &  83.29 / 73.45 & 46.64 / 41.33 & 62.92 / 60.32  \\
            \toprule[1pt]
            \bottomrule[1pt]
            \multirow{7}{*}{Waymo $\rightarrow$ Lyft} & Source Only  & 72.92 / 54.34 & 37.87 / 33.40 & 33.47 / 28.90 \\
            & SN \cite{wang2020train}  & 72.33 / 54.34 & 39.07 / 33.59 & 30.21 / 23.44 \\
            \cline{2-5}
            & ST3D  & 76.32 / {59.24} & 36.50 / 32.51 & 35.06 / 30.27 \\
            & ST3D (w/ SN) & 76.35 / 57.99 & 37.53 / 33.28 & 31.77 / 26.34 \\
            \cline{2-5}
            & ST3D++ & \textbf{79.61} / \textbf{59.93} & \textbf{40.17} / \textbf{35.47} & \textbf{37.89} / \textbf{34.49} \\
            & ST3D++ (w/ SN) & {76.67} / {58.86} & 37.89 / 34.49 & 37.73 / 32.05 \\
            \cline{2-5}
            & Oracle &  84.47 / 68.78 & 47.92 / 39.17 & 43.74 / 39.24 \\
            \toprule[1pt]
            \bottomrule[1pt]
            \multirow{7}{*}{Waymo $\rightarrow$ nuScenes} & Source Only  & 32.91 / 17.24 & 7.32 / 5.01 & 3.50 / 2.68 \\
            & SN \cite{wang2020train} & 33.23 / 18.57 & 7.29 / 5.08 & 2.48 / 1.8 \\
            \cline{2-5}
            & ST3D & {35.92} / 20.19 & 5.75 / 5.11 & {4.70} / 3.35  \\
            & ST3D (w/ SN) & 35.89 / {20.38} & 5.95 / 5.30 & 2.5 / 2.5 \\
            \cline{2-5}
            & ST3D++$^{\ddagger}$ & 35.73 / 20.90 & 12.19 / 8.91 & \textbf{5.79} / \textbf{4.84} \\
            & ST3D++ (w/ SN)$^{\ddagger}$ & \textbf{36.65} / \textbf{22.01} & \textbf{15.50} / \textbf{12.13} & 5.78 / 4.70 \\
            \cline{2-5}
            & Oracle &  51.88 / 34.87 & 25.24 / 18.92 & 15.06 / 11.73 \\
            \toprule[1pt]
            \bottomrule[1pt]
            \multirow{7}{*}{nuScenes $\rightarrow$ KITTI} & Source Only & 51.84 / 17.92 & 39.95 / 34.57 & 17.70 / 11.08 \\
            & SN \cite{wang2020train} & 40.03 / 21.23 & 38.91 / 34.36 & 11.11 / 5.67 \\
            \cline{2-5}
            & ST3D  & 75.94 / 54.13 & 44.00 / 42.60 & 29.58 / 21.21 \\
            & ST3D (w/ SN) &  {79.02} / {62.55} & 43.12 / 40.54 & 16.60 / 11.33 \\
            \cline{2-5}
            & ST3D++  & \textbf{80.52} / 62.37 & 47.20 / 43.96 & \textbf{30.87} / \textbf{23.93}  \\
            & ST3D++ (w/ SN) & 78.87 / \textbf{65.56} & \textbf{47.94} / \textbf{45.57} & 13.57 / 12.64 \\
            \cline{2-5}
            & Oracle & 83.29 / 73.45 & 46.64 / 41.33 & 62.92 / 60.32  \\
            \toprule[0.8pt]
        \end{tabular}
    }
    \end{small}
    \label{tab:SOTAcomparison}
\end{table*}

\vspace{0.1cm}
\noindent
\textbf{Adaptation Benchmark.}~
We design experiments to cover most practical 3D domain adaptation scenarios: $(i)$ Adaptation from label rich domains to label insufficient domains, $(ii)$ Adaptation across domains with different data collection locations and time (\eg Waymo $\rightarrow$ KITTI, nuScenes $\rightarrow$ KITTI), and $(iii)$ Adaptation across domains with 
a different number of the LiDAR beams (\ie Waymo $\rightarrow$ nuScenes and nuScenes $\rightarrow$ KITTI). 
Therefore, we evaluate domain adaptive 3D object detection models on the following four adaptation tasks: Waymo $\rightarrow$ KITTI, Waymo $\rightarrow$ Lyft, Waymo $\rightarrow$ nuScenes and nuScenes $\rightarrow$ KITTI. 
We rule out some ill-posed settings that are not suitable for evaluation. For example, we do not consider KITTI and Lyft as the source domain since KITTI lacks ring view annotations (less practical) and Lyft uses very different annotation rules (\ie, a large number of objects outside the road are not annotated). 
Besides, we only include most typical tasks to make the evaluation computationally manageable for follow-up research. 



\vspace{0.1cm}
\noindent
\textbf{Comparison Methods.}~
We compare ST3D++ with three methods: $(i)$ \textbf{Source Only} indicates directly evaluating the source domain pre-trained model on the target domain; $(ii)$ \textbf{SN} \cite{wang2020train} is the pioneer weakly-supervised domain adaptation method on 3D object detection with target domain statistical object size as extra information; $(iii)$ \textbf{ST3D}~\cite{yang2021st3d} is the state-of-the-art method of both unsupervised and weakly-supervised (\ie with extra target object size statistics) domain adaptation on 3D object detection; and $(iv)$ \textbf{Oracle} indicates the fully supervised model trained on the target domain.

\noindent
\textbf{Evaluation Metric.}~
We follow \cite{wang2020train} and adopt the KITTI evaluation metric for evaluating our methods on the common categories \textit{car} (also named \textit{vehicle} for similar categories in the Waymo Open Dataset) \textit{pedestrians} and \textit{cyclists} (also named \textit{bicyclist} and \textit{motorcyclist} in nuScenes and Lyft). 
Except the KITTI dataset which only provides the annotations in the front view, we evaluate the methods on ring view point clouds since they are more widely used in real-world applications, 
We follow the official KITTI evaluation metric and report the average precision (AP) in both the bird's eye view (BEV) and 3D over $40$ recall positions. The mean average precision is evaluated with IoU threshold $0.7$ for \textit{cars} and $0.5$ for \textit{pedestrians} and \textit{cyclists}. 

\noindent 
\textbf{Implementation Details.}~
We validate our proposed ST3D++ on three detection backbones SECOND \cite{yan2018second}, PointRCNN~\cite{shi2019pointrcnn} and PV-RCNN~\cite{shi2020pv}. Specifically, we improve the SECOND detector with an extra IoU head to estimate the IoU between the object proposals and their GTs, and name this detector as SECOND-IoU. Given object proposals from the RPN head in original SECOND network, we extract proposal features from 2D BEV features using the rotated RoI-align operation~\cite{he2017mask}. Then, taking the extracted features as inputs, we adopt two fully connected layers with ReLU nonlinearity~\cite{agarap2018deep} and batch normalization~\cite{ioffe2015batch} to regress the IoU between RoIs and their corresponding GTs (or pseudo boxes) with sigmoid nonlinearity. During training, we do not back-propagate the gradient from our IoU head to our backbone network.

We adopt the training settings of the popular point cloud detection codebase OpenPCDet \cite{openpcdet2020} to pre-train our detectors on the source domain with our proposed random object scaling (ROS) data augmentation strategy with scaling range $[0.7m, 1.1m]$. For the triplet box partition in hybrid quality-aware triplet memory, two thresholds $T_{\text{pos}}$ and $T_{\text{neg}}$ are typically set as 0.6 and 0.25, respectively. 
For the following target domain self-training stage, we use Adam \cite{kingma2014adam} with learning rate $1.5 \times 10^{-3}$ and one cycle scheduler to finetune the detectors for 30 epochs. We update the pseudo label with memory ensemble and voting after every 2 epochs.
For all the above datasets, the detection range is set to $[-75.2m, 75.2m]$ for $X$ and $Y$ axes, and $[-2m, 4m]$  for $Z$ axis (the origins of coordinates of different datasets have been shifted to the ground plane). 
We set the voxel size of the detector to $(0.1m, 0.1m, 0.15m)$ on all datasets. 
More detailed parameter setups could be found in our released code.

During both the pre-training and self-training processes, we adopt the widely used data augmentation, including random world flipping, random world scaling, random world rotation, random object scaling and random object rotation. CDA is utilized in the self-training process to provide proper hard examples for promoting the training process. All experiments are accomplished on 8 NVIDIA GTX 1080 TI GPUs.

\subsection{Main results and Analysis}
\label{main_results}
As shown in Table~\ref{tab:SOTAcomparison}, we compare the performance of our ST3D++ with Source Only, SN~\cite{wang2020train}, ST3D~\cite{yang2021st3d} and Oracle on four adaptation tasks. 
Since SN, one of the baseline methods, employs extra statistical supervision on the target domain, we construct our experiments of ST3D++ on two settings: one is the unsupervised DA setting including source only, ST3D and ST3D++, and the other is weakly-supervised DA setting including SN, ST3D (w/ SN) as well as ST3D++ (w/ SN), where the weakly-supervised DA setting utilizes the target object size distribution as prior. In addition, our analysis mainly focuses on two types of domain gaps as mentioned in Sec.~\ref{sec:exp_setup}.

For the content gap caused by different data-capture locations and time, the representative adaptation tasks are Waymo $\rightarrow$ KITTI and nuScenes $\rightarrow$ KITTI. For both tasks, our ST3D++ outperforms the source only and SN baseline on both UDA and weakly-supervised DA settings. Specifically, without leveraging the target domain size prior, we improve the performance on Waymo $\rightarrow$ KITTI and nuScenes $\rightarrow$ KITTI tasks by a large margin of around 38\% $\sim$ 44\%, 10\% $\sim$ 11\% and 9\% $\sim$ 10\% on car, pedestrian and cyclist separately in terms of $\text{AP}_{\text{3D}}$, which largely close the performance gap between source only and oracle. Even compared with current SOTA method ST3D~\cite{yang2021st3d}, our ST3D++ still demonstrates its superiority especially in challenging categories, e.g. pedestrian and cyclist. On Waymo $\rightarrow$ KITTI, ST3D++ exceeds ST3D by $5\%$ and $7\%$ in pedestrian and cyclist, respectively in terms of $\text{AP}_{\text{3D}}$. Besides, without bells and whistles, our ST3D++ even surpasses the oracle in pedestrian on both adaptation tasks, demonstrating the effectiveness of  ST3D++ for UDA on 3D object detection. 

In addition, since the object size gap is large between KITTI (captured in Germany) and other three datasets (all or partially captured in USA)~\cite{wang2020train}, by incorporating target object size as a prior, SN, ST3D (w/ SN) and ST3D++ (/w SN) performs prominently better in comparison with their unsupervised counterparts without SN. Especially, ST3D++ (w/ SN) even outperforms the oracle results on all evaluated categories, {\ie} car, pedestrian and cyclist on  the Waymo $\rightarrow$ KITTI task.
However, it is noteworthy that for the adaptation tasks with minor domain shifts in object size (\ie Waymo $\rightarrow$ nuScenes and Waymo $\rightarrow$ Lyft), only minor performance gains or even performance degradation are observed for SN. In contrast, our ST3D++ still obtains consistent improvements on Waymo $\rightarrow$ Lyft\footnote{Lyft dataset is constructed with different label rules from the other three datasets (\ie a large number of object that are outside the road are not annotated) which enlarges the domain gaps.}, where data for both domains are captured at similar locations with similar object distributions. 


For the point distribution gap owing to different LiDAR types, we select nuScenes $\rightarrow$ KITTI and Waymo $\rightarrow$ KITTI as representatives since they use different LiDAR types with various LiDAR beam ways (see Table~\ref{tab:dataset_infos}). When the model is adapted from a sparse domain towards a dense domain such as nuScenes $\rightarrow$ KITTI, even though the performance of the baseline model is relatively low, our ST3D++ obtains significant performance gains, {\ie} $44.45\%$, $9.39\%$ and $12.85\%$ on car, pedestrian and cyclist separately in terms of $\text{AP}_{\text{3D}}$. 
These performance gains demonstrate the advancement of our ST3D++ to improve model's capability upon a weak pre-trained detector on the source domain. 
However, when we adapt a model obtained in a dense domain to a sparse domain such as Waymo $\rightarrow$ nuScenes, the performance gains  are relatively small. In this regard, self-training strategies have more advantages on sparse to dense adaptation tasks instead of dense to sparse adaptation tasks.
The reason is that the 3D object detector trained on dense point clouds tends to make predictions with low confidence in sparse regions.
As a result, when the detector is adapted to a sparse domain, it can not generate enough high-quality pseudo labels to provide sufficient knowledge in the self-training stage.



\subsection{Comparison to Contemporary Works}

\begin{table}[htpb]
\renewcommand\arraystretch{1.1}
	\centering
	\caption{Unsupervised adaptation results of SF-UDA$^{3D}$~\cite{saltori2020sf}, Dreaming~\cite{you2021exploiting}, MLC-Net~\cite{luo2021consistency} and our ST3D++. We report $\text{AP}_{\text{3D}}$ of car at IoU 0.7 and 0.5 on nuScenes $\rightarrow$ KITTI.}
	\scalebox{0.95}{
		\begin{small}
			\begin{tabular}{l|c|c|c|c}
				\bottomrule[1pt]
				Method  & Architecture & Sequence & IoU 0.7 & IoU 0.5 \\
				\hline
				Source Only & \multirow{2}{*}{PointRCNN} & unknown &  21.9  & - \\
				SF-UDA$^{3D}$~\cite{saltori2020sf} &  & $\surd$ & 54.5  & - \\
				\hline
				Source Only & \multirow{2}{*}{PointRCNN} & unknown & - & 45.5 \\
				Dreaming~\cite{you2021exploiting} &  & $\surd$ & - & 70.3 \\
				\hline
				Source Only & \multirow{2}{*}{PointRCNN} & $\times$ & 39.55 & - \\
				MLC-Net~\cite{luo2021consistency} &  & $\times$ & 55.42 & - \\
				\hline
				Source Only & \multirow{2}{*}{SECOND-IoU} & $\times$  & 17.92  & 79.18 \\
				ST3D++ (Ours) &  & $\times$ &  {62.37} &  \textbf{87.05} \\
				\hline
				Source Only & \multirow{2}{*}{PointRCNN} & $\times$  & 42.75  & 79.24 \\
				ST3D++ (Ours) &  & $\times$ &  \textbf{67.51} &  {79.93} \\
				\toprule[0.8pt]
			\end{tabular}
		\end{small}
	}
	\label{tab:SOTA_con}
\end{table}

To further demonstrate the advancement of our ST3D++, we compare it with SF-UDA$^{3D}$~\cite{saltori2020sf}, Dreaming~\cite{you2021exploiting} and MLC-Net~\cite{luo2021consistency} on nuScenes $\rightarrow$ KITTI (\ie the only common adaptation task attempted by four approaches). 
As shown in the Table~\ref{tab:SOTA_con}, both SF-UDA$^{3D}$ and Dreaming utilize the temporal information from the point cloud sequence and an extra object tracker. Nevertheless, by only taking the single frame point cloud as input, based on SECOND-IoU, our ST3D++ outperforms SF-UDA$^{3D}$ and Dreaming by 7.87\% in $\text{AP}_{\text{3D}}$ at IoU 0.7 and 16.75\% at IoU 0.5, respectively.
Besides, ST3D++ with SECOND-IoU surpasses MLC-Net with 6.95\% in $\text{AP}_{\text{3D}}$ at IoU 0.7 with a lower baseline.

To further exclude the influence of different detection architectures, we further verify our ST3D++ on PointRCNN~\cite{shi2019pointrcnn}. 
Based on PointRCNN, our implementation for source only is 20.85\%, 33.68\% and 3.20\% stronger than the implementation by SF-UDA$^{3D}$, Dreaming and MLC-Net separately in terms of $\text{AP}_{\text{3D}}$. 
Besides, our ST3D++ with PointRCNN even achieves 67.61\% $\text{AP}_{\text{3D}}$ at IoU 0.7 thanks to the prominent two-stage refinement ability of PointRCNN, while ST3D++ based on SECOND-IoU performs more notably at IoU 0.5. 



\begin{table}[h]
\renewcommand\arraystretch{1.25}
    \centering
    \caption{Attempt of leveraging sequential point cloud information on the nuScenes $\rightarrow$ KITTI task based on SECOND-IoU. We report average precision (AP) in bird's-eye view ($\text{AP}_{\text{BEV}}$) and 3D ($\text{AP}_{\text{3D}}$) of car, pedestrian and cyclist. We indicate the best adaptation result by \textbf{bold}.}
        \begin{tabular}{c|c|c|c|c}
            \bottomrule[1pt]
            \#Frames & {Method}  & Car & {Pedestrian} & {Cyclist}\\
            \hline
            \multirow{3}{*}{1} & Source Only & 51.84 / 17.92 & 39.95 / 34.57 & 17.70 / 11.08\\
            \cline{2-5}
             & ST3D  & 75.94 / 54.13 & 44.00 / 42.60 & 29.58 / 21.21  \\
            \cline{2-5}
            & ST3D++ & {80.52} / 62.37 & \textbf{47.20} / \textbf{43.96} & {30.87} / {23.93}\\
            \hline
             \multirow{3}{*}{5} & Source Only & 62.82 / 32.08 &   29.50 / 24.66 & 20.05 / 12.07  \\
            \cline{2-5}
            & ST3D & 81.06 / 66.98 & 34.65 / 31.76 & 27.32 / 20.52 \\
            \cline{2-5}
            & ST3D++ & \textbf{80.91} / \textbf{68.23} & {30.48} / 27.86 & \textbf{29.88} / \textbf{25.57} \\
            \hline
            - & Oracle & 83.29 / 73.45 & 46.64 / 41.33 & 62.92 / 60.32 \\
            \toprule[0.8pt]
        \end{tabular}
    \label{tab:abl_multisweeps}
\end{table}

\noindent
{\bf Equip ST3D++ with temporal information.}~
Considering that contemporary works SF-UDA$^{3D}$~\cite{saltori2020sf} and Dreaming~\cite{you2021exploiting} are designed to leverage temporal information of point cloud sequences through consistency regularization, here we simply fuse several sequential point cloud frames to exploit temporal information in a straightforward manner. Furthermore, as shown in Table~\ref{tab:dataset_infos}, the point density of KITTI is around 5 times larger than nuScenes, which will cause a serious point cloud distribution gap, so our multiple frames fusion attempts can also mitigate point density gaps. Specifically, here we merge five LiDAR pose calibrated frames in nuScenes to approach the point cloud density in KITTI. We extend the point cloud with an extra timestamp channel to identify different frames in nuScenes. For the target domain KITTI, we just pad zeros to each point as the timestamp channel.

As shown in Table~\ref{tab:abl_multisweeps}, for the source only model, directly adapting the model from densified nuScenes to KITTI brings around 14\% and 1\% gains on car and cyclist separately in terms of $\text{AP}_{\text{3D}}$. Besides, with multi-frame fused source data, ST3D++ obtains around 5.9\% and 1.6\% improvements in $\text{AP}_{\text{3D}}$ and achieves new SOTA performance on car and cyclist. These experimental results demonstrate that our ST3D++ can largely benefit from temporal information even in a simple fusion manner. We believe that ST3D++ can consistently benefit from the development of temporal-based 3D detection from point clouds 
since they are orthogonal. 
Note that pedestrian suffers from performance degradation with five frames fused source data. The reason might lie in the different characteristics of rigid and nonrigid categories. Gestures of nonrigid pedestrians are very diverse along different frames, while the same car on different frames has same shape as rigid object. As a result, directly fusing multiple frames might produce nondescript gestures for nonrigid categories and finally confuse the deep learner.

\subsection{ST3D++ Results with SOTA Detection Architecture}
\begin{table}[htpb]
\renewcommand\arraystretch{1.1}
	\centering
	\caption{Adaptation results based on the state-of-the-art detector PV-RCNN~\cite{shi2020pv}. We report $\text{AP}_{\text{BEV}}$ / $\text{AP}_{\text{3D}}$ of car, pedestrian and cyclist on Waymo $\rightarrow$ KITTI.}
		\begin{small}
			\begin{tabular}{l|c|c|c}
				\bottomrule[1pt]
				Method & Car & Pedestrian & Cyclist \\
				\hline
				Source Only & 61.18 / 22.01 & 46.65 / 43.18 & 54.40 / 50.56  \\
				SN & 79.78 / 63.60 & 54.78 / 53.04 & 52.65 / 49.56 \\
				\hline
				ST3D & 84.10 / 64.78 & 50.15 / 47.24 & 51.63 / 48.23 \\
				ST3D (w/ SN) & 86.65 / 76.86 & 55.23 / 53.20 & 56.84 / 53.71 \\
				\hline
				ST3D++ & {84.59} / {67.73} & 56.63 / 53.36 &  58.64 / 55.07 \\
				ST3D++ (w/ SN) & \textbf{86.92} / \textbf{77.36} &  \textbf{63.58} / \textbf{62.87} & \textbf{65.61} / \textbf{61.42} \\
				\hline
				Oracle & 88.98 / 82.50 & 54.13 / 49.96 & 73.65 / 70.69 \\
				\toprule[0.8pt]
			\end{tabular}
		\end{small}
	\label{tab:pv_results}
\end{table}
We further investigate the generalization of our ST3D++ by employing more sophisticated detection architecture PV-RCNN~\cite{shi2020pv} as the base detector without any specific hyper-parameter adjustment. As shown in Table~\ref{tab:pv_results}, for the adaptation task Waymo $\rightarrow$ KITTI,
our unsupervised ST3D++ outperforms source only by 45.72\%, 10.18\% and 4.51\% on car, pedestrian and cyclist separately in terms of AP$_{\text{3D}}$. It also surpasses the SOTA cross-domain 3D object detection method ST3D with 2.95\%, 6.12\% and 6.84\% on car, pedestrian and cyclist in AP$_{\text{3D}}$. Furthermore, incorporated with SN, our ST3D++ (w/ SN) is further improved to approach oracle results since SN provides better localized pseudo labels for subsequent self-training with the prior of target object statistics. These prominent experimental results strongly demonstrate that our ST3D++ is a model-agnostic self-training pipeline. 
Our ST3D++ can further harvest the progress of 3D object detector through the produced more accurate pseudo labels.

\section{Ablation Studies}
\subsection{Component Analysis of ST3D++}
\label{sec:abl_study}
In this section, we conduct extensive ablation experiments to investigate the individual components of our ST3D++. All experiments are conducted on  SECOND-IoU for the adaptation task of Waymo $\rightarrow$ KITTI.
Notice for category-agnostic components such as memory updating and curriculum data augmentation, the ablation experiments are only conducted on the car category.  

\begin{table}[h]
\renewcommand\arraystretch{1.1}
    \centering
    \caption{Effectiveness analysis of Random Object Scaling in terms of AP$_{\text{BEV}}$ / AP$_{\text{3D}}$ on car, pedestrian and cyclist.}
        \begin{tabular}{l|c|c|c}
            \bottomrule[1pt]
            Method  & Car & Pedestrian & Cyclist \\
            \hline
            (a) Source Only & 67.64 / 27.48 & 46.29 / 43.13 & 48.61 / 43.84 \\ 
            (b) ROS &  78.07 / 54.67 & 49.90 / 46.43 & 50.61 / 46.96 \\
            (c) SN  & 78.96 / 59.20 & 53.72 / 50.44 & 44.61 / 41.43 \\
            \hline
            (d) ST3D++ (w/o ROS) & 77.35 / 33.73 & 52.03 / 48.13 &  57.83 / 50.81 \\
            (e) ST3D++ (w/ ROS)  &  80.78 / 65.64 & 55.98 / 53.30 & 57.88 / 52.87 \\
            (f) ST3D++ (w/ SN) & \textbf{86.47} / \textbf{74.61} & \textbf{52.10} / \textbf{59.21} & \textbf{65.07} / \textbf{60.76} \\
            \toprule[0.8pt]
        \end{tabular}
    
    \label{tab:random_object_size}
\end{table}

\begin{table*}[ht!]
\renewcommand\arraystretch{1.1}
    \centering
    \caption{Component ablation studies on car, pedestrian and cyclist in terms of AP$_{\text{BEV}}$ / AP$_{\text{3D}}$. \textbf{ST} represents naive self-training. \textbf{Triplet} means the triplet box partition.
    \textbf{QAC} is to use the IoU as a criterion without considering the classification confidence as in ST3D~\cite{yang2021st3d}.
    \textbf{HQAC} indicates the hybrid quality-aware criterion. \textbf{MEV-C} is memory ensemble and memory voting based on consistency. \textbf{CDA} means curriculum data augmentation. \textbf{SASD} stands for source-assisted self-denoised training.}
    \begin{small}
    	\setlength\tabcolsep{7pt}
        \begin{tabular}{l|c|c|c|c|c|c|c|c|c}
            \bottomrule[1pt]
            ST & Triplet & MEV-C & QAC & HQAC & CDA & SASD & Car & Pedestrian & Cyclist \\
            \hline
             & & & & & & & 78.96 / 59.20 & 53.72 / 50.44 & 44.61 / 41.43 \\
            \hline
            $\surd$ & & & & & & & 79.74 / 65.88 & 51.18 / 49.13 & 51.65 / 48.61 \\
            $\surd$ & $\surd$ & & & & & &  79.81 / 67.39 & 52.72 / 49.73 & 54.29 / 50.33 \\
            $\surd$ & $\surd$ & $\surd$ & & & & & 82.72 / 70.17  & 54.06 / 51.13 &  54.72 / 52.85 \\
            $\surd$ & $\surd$ & $\surd$ & $\surd$ & & & & 85.35 / 72.52 & 54.74 / 51.92 & 56.19 / 53.00 \\
            $\surd$ & $\surd$ & $\surd$ & & $\surd$ & & & 85.35 / 72.52 & 59.36 / 56.13 & 57.38 / 53.55 \\
            $\surd$ & $\surd$ & $\surd$ & & $\surd$ & $\surd$ & &  {85.83} / {73.37} & 59.97 / 56.27 & 56.30 / 53.49 \\
            $\surd$ & $\surd$ & $\surd$ & & $\surd$ & $\surd$ & $\surd$ & \textbf{86.47} / \textbf{74.61} & \textbf{62.10} / \textbf{59.21} & \textbf{65.07} / \textbf{60.76} \\
            \toprule[0.8pt]
        \end{tabular}
    \end{small}
    \label{tab:component_analysis}
\end{table*}

\vspace{0.1cm}
\noindent
\textbf{Random Object Scaling.}~
Here we investigate the effectiveness of our unsupervised random object scaling (ROS) for mitigating the domain shift of object size statistics across domains as mentioned in Sec.~\ref{sec:pretrain}.
By employing random object scaling as one of the data augmentations for pre-training, the detector could be more robust to variations of object sizes in different domains.
As shown in Table~\ref{tab:random_object_size} (a), (b), (c), our unsupervised ROS improves the performance by around 27.2\%, 3.3\% and 3.1\% on car, pedestrian and cyclist respectively in terms of $\text{AP}_\text{3D}$. Besides, our ROS is only 4.5\% lower than the weakly-supervised SN method on car and even surpasses SN on cyclist.
Furthermore, as shown in Table~\ref{tab:random_object_size} (d), (e), 
the ROS pre-trained model also greatly benefits the subsequent self-training process since it provides pseudo labels with less noise.
We also observe that there still exists a gap between the performance of ST3D++ (w/ ROS) and ST3D++ (w/ SN) in $\text{AP}_\text{3D}$, potentially due to that the KITTI dataset has a larger domain gap over object sizes compared with other datasets, and under this situation, the weakly-supervised SN could leverage accurate object size information and  generate pseudo labels with less localization noise.

\vspace{0.1cm}
\noindent
\textbf{Component Analysis in Self-training.}~
As demonstrated in Table~\ref{tab:component_analysis}, we investigate the effectiveness of our individual components at the self-training stage. Our ST3D++ (w/ SN) (last line) outperforms the SN baseline and naive ST baseline by 15.14\% and 8.73\%  on car, 16.08\% and 10.08\% on pedestrian as well as 11.52\% and 6.35\% on cyclist in terms of $\text{AP}_{\text{3D}}$, manifesting the effectiveness of self-training on domain adaptive 3D object detection. 

In the pseudo label generation stage, hybrid quality-aware triplet memory (HQTM), including hybrid quality-aware criterion, triplet box partition and memory ensemble-and-voting, is designed to address pseudo label noise, and yield an improvement of 6.6\%, 7\%, and 4.9\%  in $\text{AP}_{\text{3D}}$ on car, pedestrian and cyclist respectively (See Table~\ref{tab:component_analysis}). 
First, by avoiding assigning labels to ambiguous samples, the triplet boxes partition scheme brings 1.51\%, 0.6\% and 2.56\% improvements on car, pedestrian and cyclist in $\text{AP}_{\text{3D}}$. 
Then, Memory ensemble and voting integrates historical pseudo labels and stabilizes the pseudo label updating process, leading to an improvement of around 1.5\% $\sim$ 2.8\% in $\text{AP}_{\text{3D}}$ on all evaluated categories. 
Finally, the quality-aware criterion in \cite{yang2021st3d} which only considers the IoU as a criterion, has improved the performance of car by 2.35\%, while boosting the performance of other categories, {\ie} pedestrian and cyclist only slightly. In contrast, our hybrid quality-aware criterion consistently boosts the evaluated categories car, pedestrian and cyclist by 2.35\%, 4.21\% and 0.7\% . The above shows HQAC is instrumental in improving performance.  

In the model training stage,  source-assisted self-denoised training (SASD) and curriculum data augmentation (CDA) jointly deliver around 2.1\%, 3.1\% and 7.2\% improvements in $\text{AP}_{\text{3D}}$ on car, pedestrian and cyclist respectively. 
In particular, small categories such as pedestrian and cyclist benefit more from SASD since they typically suffer from more severe classification and localization noise which will mislead the optimization process while SASD efficiently corrects the direction of gradient descent in the training stage and significantly improves the performance, \ie 1.24\%, 2.94\%, and 7.27\% for car, pedestrian and cyclist in $\text{AP}_{\text{3D}}$ respectively.

\vspace{0.1cm}
\noindent
\textbf{Analysis of Memory Ensemble and Voting.}
\label{abl:mev}
Here, we further investigate the memory ensemble and memory voting schemes for memory updating and consistent pseudo label generation.
The analysis contains three aspects: different memory ensemble strategies, the advancement of memory voting and pseudo label merging methods in memory ensemble. The comparison results are shown in Table~\ref{tab:abl_memory_ensemble}.
First, for the comparison of different memory ensemble variants, three variations (see Sec.~\ref{sec:memoryensemble} for details) achieve similar performance, and bipartite ensemble outperforms 1.4\% and 0.6\% than ME-N and ME-C respectively in terms of $\text{AP}_{\text{3D}}$.
For the paired box merging methods (see Fig.~\ref{fig:memory_ensemble}), we compare two merging approaches ``max score'' and ``weighted average'', where max score obtains a 2.5\% performance gain than the weighted average.  
This validates our analysis in Sec.~\ref{sec:memoryensemble} that simply selecting box with higher confidence can generate better pseudo labels.
In addition, without memory voting, the performance of ST3D++ (w/ ME-C) drops by around 1.9\% in terms of $\text{AP}_{\text{3D}}$ since the unmatched boxes along different memories could not be properly handled. 
Our memory voting strategy could robustly mine high-quality boxes and discard low-quality boxes.

\begin{table}
\renewcommand\arraystretch{1.1}
	\centering
	\caption{Investigations of memory updating strategies for pseudo labels on car in terms of AP$_{\text{BEV}}$ / AP$_{\text{3D}}$. We consider different variants of memory ensemble, different manners for merging matched pseudo labels and proxy-pseudo labels and the effectiveness of memory voting. Three types of memory ensemble variants: consistency, NMS and bipartite are denoted as ME-C, ME-N, ME-B separately.}
	\scalebox{0.98}{
		\setlength\tabcolsep{1pt}
		\begin{small}
			\begin{tabular}{c|c|c|c}
				\bottomrule[1pt]
				Method  & Merge Manner & Memory Voting   & Car \\
				\hline
				ST3D++ (w/ ME-N) &  Max & $\surd$ & 86.69 / 74.21 \\
				ST3D++ (w/ ME-B) &  Max & $\surd$ & \textbf{86.84} / \textbf{75.22} \\
				\hline
				\multirow{4}{*}{ST3D++ (w/ ME-C)} & Max & $\surd$ & 86.47 / 74.61 \\
				& Avg & $\surd$ & 82.35 / 72.09 \\
				\cline{2-4}
				& Max & $\times$ & 86.41 / 72.76 \\
				& Avg & $\times$ & 86.60 / 72.32 \\ 
				\toprule[0.8pt]
			\end{tabular}
		\end{small}
	}
	
	\label{tab:abl_memory_ensemble}
\end{table}

\begin{table}[h]
\renewcommand\arraystretch{1.1}
    \centering
    \caption{Analysis of data augmentation type and intensity on car in terms of AP$_{\text{BEV}}$ / AP$_{\text{3D}}$.}
    \scalebox{0.98}{
    \begin{small}
        \begin{tabular}{c|cc|c|c}
            \bottomrule[1pt]
            Method  & World & Object & Intensity & Car \\
            \hline
            \multirow{6}{*}{ST3D++ (w/ SN)} &$\times$  &$\times$  & - & 84.62 / 69.62 \\
             & $\surd$ & $\times$ & Normal & 84.17 / 69.17 \\
             &$\times$  & $\surd$ & Normal & 86.78 / 73.63 \\
             & $\surd$ & $\surd$ & Normal & 86.65 / 73.98  \\
            \cline{2-5} 
            & $\surd$ & $\surd$ & Strong & \textbf{86.71} / 73.59 \\
            & $\surd$ & $\surd$ & Curriculum & {86.47} / \textbf{74.61} \\
            \toprule[0.8pt]
        \end{tabular}
    \end{small}
    }
    \label{tab:abl_aug}
    
\end{table}

\vspace{0.1cm}
\noindent
\textbf{Data Augmentation Analysis.}
As shown in Table~\ref{tab:abl_aug}, we also explore the influence of different data augmentation strategies and intensities in the model training stage. 
We divide all data augmentation into two groups: world-level data augmentations that affect the whole scene (\ie random world flipping, random world scaling and random world rotation) and object-level data augmentations that change each instance (\ie random object rotation and random object scaling).
We observe that without any data augmentation, ST3D++ suffers from around 5\% performance degradation. 
Object-level data augmentations provide significant improvements of around 4\% in terms of AP$_{\text{3D}}$ while world-level data augmentations even slightly harm performance.
The combination of object- and world-level data augmentations further improves the detector's capability.  
When it comes to the intensity of data augmentation (see Sec.~\ref{sec:cda}), compared to the normal intensity, simply adopting stronger data augmentation magnitude confuses the deep learner and suffers from slightly  performance drop while our CDA, progressively enlarges the intensities, can bring around 0.6\% gains.

\begin{table}[h]
\renewcommand\arraystretch{1.1}
    \centering
    \caption{Analysis of the effectiveness of DSNorm in SASD module on Waymo $\rightarrow$ KITTI in terms of AP$_{\text{BEV}}$ / AP$_{\text{3D}}$.}
    \scalebox{0.9}{
    \begin{small}
        \begin{tabular}{c|c|c|c|c}
            \bottomrule[1pt]
            SASD & DSNorm & Car & Pedestrian & Cyclist  \\
            \hline
             $\times$ & $\times$  &  85.83 / 73.37 & 59.97 / 56.27 & 56.30 / 53.49 \\
             $\surd$ & $\times$ & 84.10 / 68.08 & 58.70 / 55.96   & 59.65 / 56.88 \\
            $\surd$ & $\surd$  & \textbf{86.47} / \textbf{74.61} & \textbf{62.10}/ \textbf{59.21}  &  \textbf{65.07} / \textbf{60.76} \\
            \toprule[0.8pt]
        \end{tabular}
    \end{small}
    }
    \label{tab:abl_dsbn}
\end{table}

\begin{figure*}[h]
    \centering
    \includegraphics[width=1\linewidth]{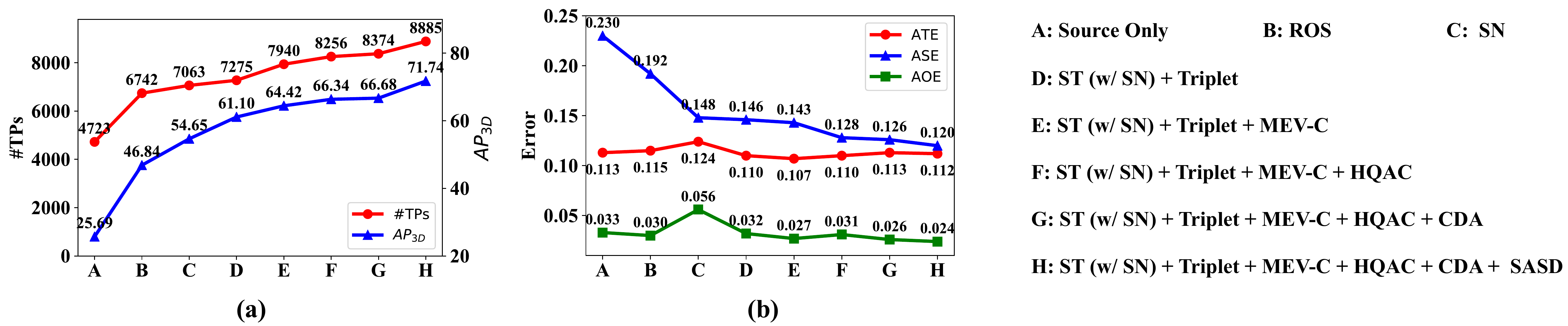}
    \caption{Quality metrics of pseudo labels for car on the KITTI training set using SECOND-IoU. \textbf{\#TPs} indicates the number of true-positive predictions at IoU $0.7$ among pseudo labels. \textbf{$AP_{\text{3D}}$} stands for the average precision in 3D over 40 recall positions. \textbf{ATE}, \textbf{ASE} and \textbf{AOE} represent average translation, scale and orientation errors of correct pseudo labels. }
    \label{fig:quality}
\end{figure*}

\vspace{0.1cm}
\noindent
\textbf{Effect of Domain-Specific Normalization in SASD.} 
As mentioned in Sec.~\ref{sec:sasd}, although the assistance of source data can help rectify the gradient directions and offer hard cases to train the model, a noteworthy issue is the distribution shift caused by statistic differences of batch normalization layers.
Here, we investigate the effectiveness of Domain-Specific Normalization (DSNorm) in addressing this issue.
As illustrated in Table~\ref{tab:abl_dsbn}, without DSNorm, the naive SASD even causes performance degradation on car and pedestrian due to domain shifts. 
After being equipped with DSNorm, SASD obtains obvious improvements particular for pedestrian and cyclist (\ie 2.94\% and 7.27\% in terms of $AP_{\text{3D}}$ separately).

\subsection{Effects of Each Component on Pseudo Label Qualities}
We investigate how each module benefits self-training by analyzing their contributions on correcting pseudo label noise  and improving pseudo label qualities.
We adopt $AP_{\text{3D}}$ and \#TPs to assess the correctness of pseudo labels, and employ \textbf{ATE}, \textbf{ASE} and \textbf{AOE} to assess the average translation, scale and orientation errors of pseudo labels. The later is inspired by nuScenes toolkit~\cite{caesar2020nuscenes}: 
Average translation error (ATE) is the euclidean object center distance in 2D from bird's eye view (measured in meters); Average scale error (ASE) is the 3D intersection over union (IoU) of prediction and its corresponding ground truth after aligning heading angle and object center (calculated by $1 - \text{IoU}$); and Average orientation error is the smallest yaw angle difference between the pseudo label and the ground truth (measured in radian).


As shown in Fig.~\ref{fig:quality}, when the proposed components are gradually incorporated to the pipeline (\ie C $\rightarrow$ H), the number of true positives and $\text{AP}_{\text{3D}}$ are progressively increased.
This tendency illustrates that the classification noise is significantly reduced and thus its correctness is improved by a large margin.
Specifically, ROS mitigates domain differences in object size distributions and hence largely reduces ASE.
With Triplet, HQAC and MEV, our method generates accurate and stable pseudo labels, localizing more \#TPs with fewer errors.
CDA overcomes overfitting and reduces both ASE and AOE.
SASD reduces ASE and AOE through addressing the negative impacts of pseudo label noise on model training.

\section{Conclusion}
We have presented ST3D++ -- a holistic denoised self-training pipeline for unsupervised domain adaptive 3D object detection from point clouds.
ST3D++ redesigns different self-training stages from model pre-training on source labeled data, pseudo label generation on target data to model training on pseudo labeled data.
At the first two stages, it effectively reduces pseudo label noise through pre-training a de-biased object detector via random object scaling and designing a robust pseudo label assignment method, a better pseudo label selection criterion and a consistency regularized pseudo label update strategy.
These components cooperate to make pseudo labels accurate and consistent. 
In addition, at the model training stage, the negative impacts of noisy pseudo labels are alleviated via the assistance of supervision from source labeled data .  Simultaneously, curriculum data augmentation is also developed to overcoming the overfitting issue on easy pseudo labeled target data.
Our extensive experimental results demonstrate that ST3D++ substantially advances state-of-the-art methods. 
Our future work would be to extend our ST3D++ to indoor scenes or sequence data and investigate point cloud translation approaches to address domain gaps in point distributions.


\bibliographystyle{IEEEtran}
\bibliography{IEEEabrv,egbib}

\end{document}